%% file: Master.tex
\documentclass[twoside,11pt]{article}

\usepackage{jmlr2e}

\usepackage{microtype}
\usepackage{graphicx}
\usepackage{subfigure}
\usepackage{booktabs} 
\usepackage{hyperref}
\usepackage{algorithm}
\usepackage{algorithmic}
\usepackage{xspace}
\usepackage{amsmath,amsfonts,epsf,psfrag,amssymb,bbm,bm}
\usepackage{setspace}

\input{symbolsdef}

\jmlrheading{1}{2018}{1-48}{4/00}{10/00}{Kamyar Azizzadenesheli, Alessandro Lazaric, and Animashree Anandkumar}

\ShortHeadings{Reinforcement Learning in Rich-Observation MDPs using Spectral Methods}{Azizzadenesheli, Lazaric, Anandkumar}
\firstpageno{1}

\begin{document}

\title{Reinforcement Learning in Rich-Observation MDPs using Spectral Methods}

\author{\name Kamyar Azizzadenesheli \email kazizzad@uci.edu\\
       \addr Department of EECS\\
       University California, Irvine\\
       Irvine, CA 92697, USA
       \AND
       \name Alessandro Lazaric \email alessandro.lazaric@inria.fr \\
       \addr Facebook AI Research (FAIR)\\
       Paris, France
       \AND
       \name Animashree Anandkumar \email anima@caltech.edu \\
       \addr California institute of technology \\
       1200 E California Blvd, Pasadena, CA 91125}

\editor{}

\maketitle

\begin{abstract}
Reinforcement learning (RL) in Markov decision processes (MDPs) with large state spaces is a challenging problem. The performance of standard RL algorithms degrades drastically with the dimensionality of state space. However, in practice, these large MDPs typically incorporate a latent or hidden low-dimensional structure. In this paper, we study the setting of {\em rich-observation} Markov decision processes (\richmdp), where there are a small number of hidden states which possess an injective mapping to the observation states. In other words, every observation state is generated through a single hidden state, and this mapping is unknown a priori.  We introduce a spectral decomposition method that consistently learns this mapping, and more importantly, achieves it with low regret. The estimated mapping is integrated into an optimistic RL algorithm (UCRL), which operates on the estimated hidden space. We derive finite-time regret bounds for our algorithm with a weak dependence on the dimensionality of the observed space. In fact, our algorithm asymptotically achieves the same average regret  as the oracle UCRL algorithm, which has the knowledge of the mapping from hidden to observed spaces. Thus, we derive an efficient spectral RL algorithm for ROMDPs.
\end{abstract}

\input{1Introduction}
\input{3Preliminaries}
\input{4RLalgorithm}
\input{4.1Experiment}

\input{5Conclusion}



\paragraph*{Acknowledgments}
K. Azizzadenesheli is supported in part by NSF Career Award CCF-1254106 and AFOSR YIP FA9550-15-1-0221. A. Lazaric is sup- ported in part by a grant from CPER Nord-Pas de Calais/FEDER DATA Advanced data science and technologies 2015-2020, CRIStAL (Centre de Recherche en Informatique et Automatique de Lille), and the French National Research Agency (ANR) under project ExTra-Learn n.ANR-14- CE24-0010-01. A. Anandkumar is supported in part by Microsoft Faculty Fellowship, Google fac- ulty award, Adobe grant, NSF Career Award CCF- 1254106, AFOSR YIP FA9550-15-1-0221, and Army Award No. W911NF-16-1-0134. The work is partially developed when the first K. Azizzadenesheli was visiting INRIA, Lille and Simons Institute for the Theory of Computing, UC. Berkeley. 
\newpage

\appendix
\input{6Supplementary}

\input{6Supplementary_regret}
\input{6.1FurtherClustering}

\input{6.01Consentrations}

\input{7MoreExperiments}

\newpage

\vskip 0.2in
\bibliography{Master}

\end{document}

%% file: symbolsdef.tex
\newcommand{\Omin}{O_{\min}}

\newcommand{\BO}{\mathcal{B}_O}

\newcommand{\wt}[1]{\widetilde{#1}}
\newcommand{\wh}[1]{\widehat{#1}}

\newcommand{\wb}[1]{\overline{#1}}

\newcommand{\Pportion}{\alpha_p}
\newcommand{\transp}{\mathsf{T}}

\newcommand{\richmdp}{ROMDP\xspace}
\newcommand{\richmdps}{ROMDPs\xspace}
\newcommand{\first}{\text{first}}
\newcommand{\last}{\text{last}}
\newcommand{\sm}{\text{sm}}

\newcommand{\vecv}{\vec v}

\newcommand{\smcmdp}{\textsc{\small{SL-UCRL}}\xspace}
\newcommand{\scucrl}{\textsc{\small{SL-UCRL}}\xspace}
\newcommand{\ucrl}{\textsc{\small{UCRL}}\xspace}
\renewcommand{\Re}{\mathbb{R}}
\renewcommand{\S}{\mathcal{S}}
\newcommand{\F}{\mathcal F}
\newcommand{\A}{\mathcal A}
\newcommand{\X}{\mathcal X}
\newcommand{\Y}{\mathcal Y}
\newcommand{\M}{\mathcal M}
\newcommand{\E}{\mathbb E}
\newcommand{\Prob}{\mathbb P}
\newtheorem{assumption}{Assumption}

%% file: 1Introduction.tex

\section{Introduction}\label{sec:intro}
Reinforcement learning (RL) framework studies the problem of efficient agent-environment interaction,
where the agent learns to maximize 
a given reward function in the long run~\citep{bertsekas1996neuro-dynamic,sutton1998introduction}. 
At the beginning of the interaction, 
the agent is uncertain about the environment's dynamics 
and must \textit{explore} different policies in order to gain information about it.
Once the agent is fairly certain, the
knowledge about the environment can be \textit{exploited}
to compute a good policy attaining a large cumulative reward. 
Designing algorithms that achieve an effective trade-off between exploration and exploitation 
is the primary goal of reinforcement learning. 
The trade-off is commonly measured 
in terms of {\em cumulative regret}, that is the difference between 
the rewards accumulated by the optimal policy 
(which requires exact knowledge of the environment) 
and those obtained by the learning algorithm.

In practice, we often deal with environments 
with large observation state spaces (e.g., robotics). In this case the regret of standard RL algorithms grows quickly with the size of the observation state space. (We use observation state and observation interchangeably.)
Nonetheless, in many domains there is an underlying low dimensional latent space that summarizes the large observation space and its dynamics and rewards. For instance,  in robot navigation, the high-dimensional visual and sensory input 
can be summarized into a 2D position map, but this map is typically unknown.
%
This makes the problem challenging, since it is not immediately clear how to exploit the low-dimensional latent structure to achieve low regret.

\textbf{Contributions.} In this paper we focus on rich-observation Markov decision processes (\richmdp), where a small number of $X$ hidden states are mapped to a large number of $Y$ observations through an injective mapping, so that an observation can be generated by only one hidden state and hidden states can be viewed as clusters.

In this setting, we show that it is indeed possible to devise an algorithm that starting from observations can progressively cluster them in ``smaller'' states and eventually converge to the hidden MDP. We introduce \smcmdp, where we integrate spectral decomposition methods into the upper-bound for RL algorithm (UCRL)~\citep{jaksch2010near}. The algorithm proceeds in epochs in which an estimated mapping between observations and hidden state is computed and an optimistic policy is computed on the MDP (called auxiliary MDP) constructed from the samples collected so far and the estimated mapping. The mapping is computed using spectral decomposition of the tensor associated to the observation process. 

We prove that this method is guaranteed to correctly ``cluster'' observations together with high probability. As a result, the dimensionality of the auxiliary MDP decreases as more observations are clustered, thus making the algorithm more efficient computationally and more effective in finding good policies. Under transparent and realistic assumptions, we derive a regret bound showing that the per-step regret decreases over epochs, and we prove a worst-case bound on the number of steps (and corresponding regret) before the full mapping between states and observations is computed. The regret accumulated over this period is actually constant as the time to correct clustering does not increase with the number of steps $N$. As a result,  \smcmdp asymptotically matches the regret of learning directly on the latent MDP. We also notice that the improvement in the regret comes with an equivalent reduction in time and space complexity. In fact, as more observations are clustered, the space to store the auxiliary MDP decreases and the complexity of the extended value iteration step in UCRL decreases from $O(Y^3)$ down to $O(X^3)$.

\textbf{Related work.}
The assumption of the existence of a latent space 
is often used to reduce the learning complexity. 
For multi-armed bandits, \citet{gheshlaghi-azar2013sequential} and \citet{maillard2014latent} assume that a bandit problem 
is generated from an unknown (latent) finite set 
and show how the regret can be significantly reduced 
by learning this set. 
\citet{gentile2014online} consider the more general scenario 
of latent contextual bandits,
where the contexts belong to a few underlying hidden classes. 
They show that a uniform exploration strategy over the contexts, combined with an online clustering algorithm 
achieve a regret scaling only with the number of hidden clusters. An extension to recommender systems is considered in~\citet{gopalan2016low} where the contexts for the users and items are unknown a priori. Again, uniform exploration is used together with the spectral algorithm of~\citet{anandkumar2014tensor} to learn the latent classes. Bartók et al. ~\citep{bartok2014partial} tackles a general case of partial monitoring games and provides minimax regret guarantee which is polynomial in certain dimensions of the problem.

The \richmdp model considered is a generalization of the latent contextual bandits, where actions influence the contexts (i.e., the states) and the objective is to maximize the long-term reward. 
\richmdps have been studied in~\citet{krishnamurthy2016pac} in the PAC-MDP setting and episodic deterministic environments using an algorithm searching the best $Q$-function in a given function space. This result is extended to the general class of contextual decision processes in~\citet{jiang2016contextual}. While the resulting algorithm is proved to achieve a PAC-complexity scaling with the number of hidden states/factors $X$,  it suffers from high computations complexity.

\citet{ortner2013adaptive} proposes an algorithm integrating state aggregation with UCRL but, while the resulting algorithm may significantly reduce the computational complexity of UCRL, 
the analysis does not show any improvement in the regret. 

Learning in \richmdps can be also seen as a state-aggregation problem, where observations are aggregated to form a small latent MDP. While the literature on state-aggregation in RL is vast, most of the results have been derived for the batch setting (see e.g.,~\citet{li2006towards}).  

Finally, we notice that \richmdps are a special class of partially observable MDPs (POMDP). \citet{azizzadenesheli2016reinforcement} recently proposed an algorithm that leverages spectral methods to learn the hidden dynamic of POMDPs and derived a regret scaling as $\sqrt{Y}$ using fully stochastic policies (which are sub-optimal in \richmdps). While the computation of the optimal memoryless policy relies on an optimization oracle, which in general is NP-hard~\citet{littman1994memoryless,vlassis2012computational,porta2006point,azizzadenesheli2016open,shani2013survey}, computing the optimal policy in \richmdps amounts to solving a standard MDP. Moreover, \citet{guo2016pac} develops a PAC-MDP analysis for learning in episodic POMDPs and obtain a bound that depends on the size of the observations. The planning, in general, is computationally hard since it is a mapping from history to action.

%% file: 3Preliminaries.tex

\section{Rich Observation MDPs}

\begin{figure}[t!]
\hspace{-0.2in}
  \centering
  \begin{psfrags}
\psfrag{x1}[][1]{$x_{t}$}
\psfrag{x0}[][1]{$x_t$}
\psfrag{x2}[][1]{$x_{t+1}$}
\psfrag{x3}[][1]{$x_{t+2}$}
\psfrag{y0}[][1]{$y_t$}
\psfrag{y1}[][1]{$y_{t}$}
\psfrag{y2}[][1]{$y_{t+1}$}
\psfrag{r0}[][1]{$r_t$}
\psfrag{r1}[][1]{$r_{t}$}
\psfrag{r2}[][1]{$r_{t+1}$}
\psfrag{a0}[][1]{$a_{t}$}
\psfrag{a1}[][1]{$a_{t}$}
\psfrag{a2}[][1]{$a_{t+1}$}
\includegraphics[scale=0.20]{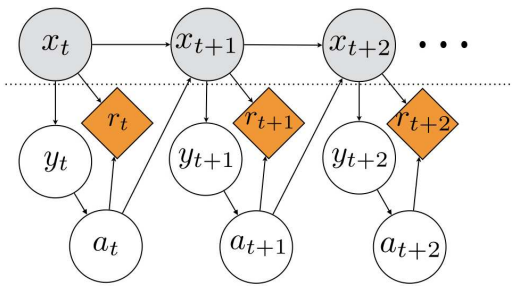}
\end{psfrags}
\caption{Graphical model of a \richmdp.}
\label{fig:Gmodel}
\end{figure}

\begin{figure}[ht]
\centering
  \includegraphics[height=4.cm]{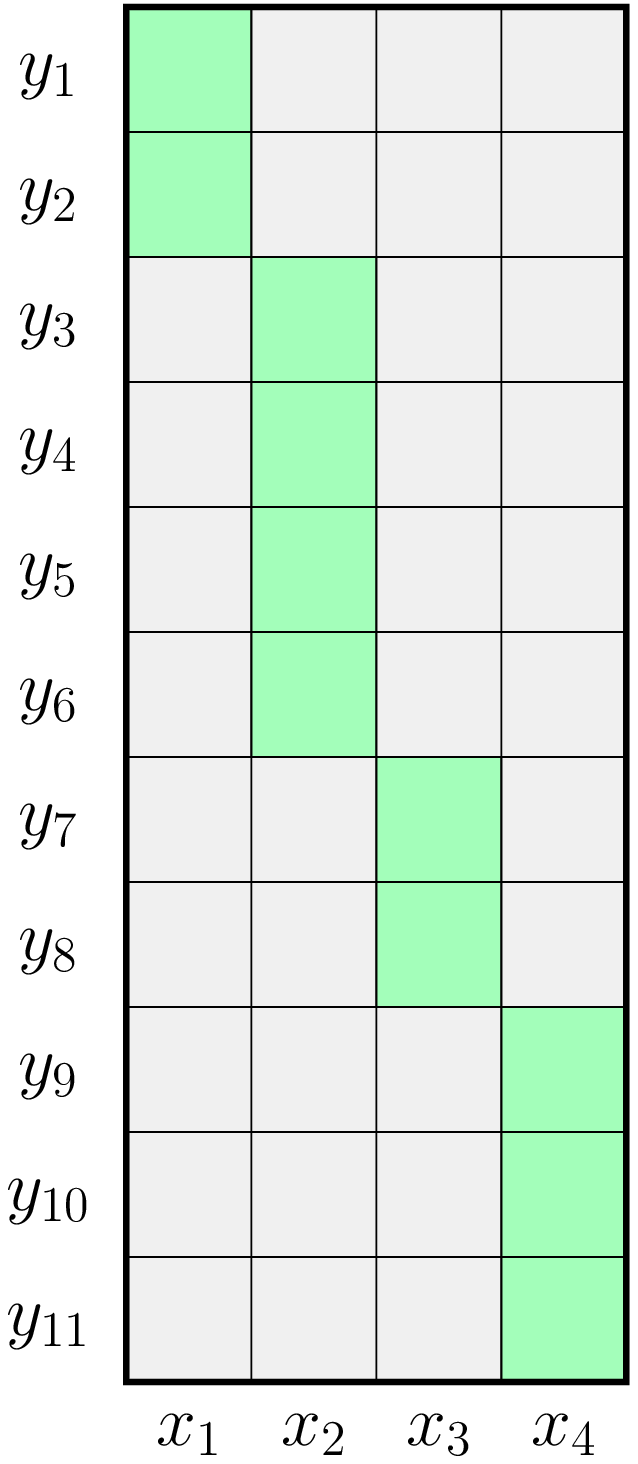}
\hspace{0.5in}
  \centering
  \includegraphics[height=4.cm]{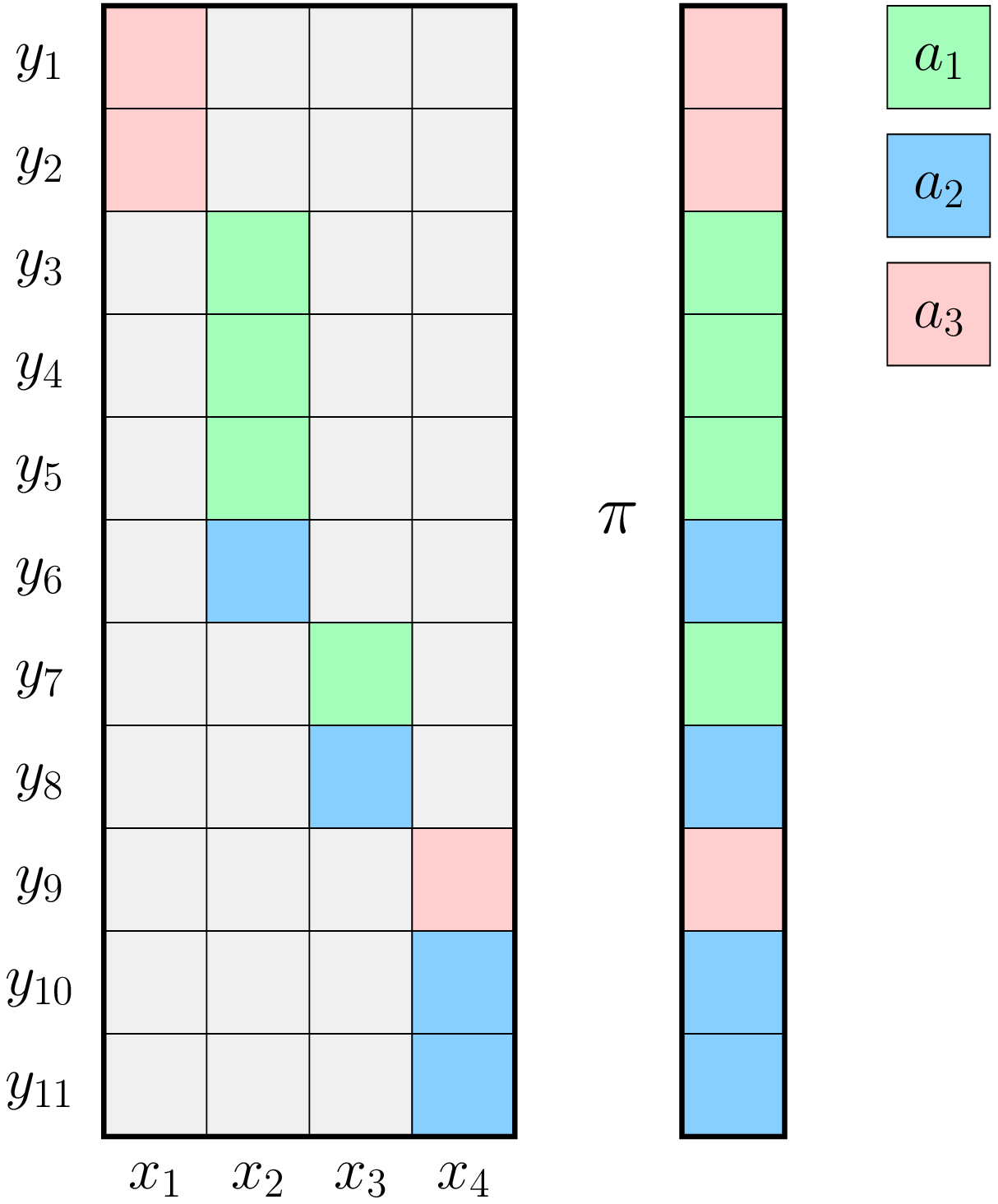}
\caption{\textit{(left)} Example of an observation matrix $O$. Since state and observation labeling is arbitrary, we arranged the non-zero values so as to display a diagonal structure. \textit{(right)} Example of clustering that can be achieved by policy $\pi$ (e.g., $\X_{\pi}^{(a_1)} = \{x_2, x_3\}$). Using each action we can recover \textit{partial} clusterings corresponding to 7 auxiliary states $\S =\{s_1 .. s_7\}$ with clusters $\Y_{s_1} = \{y_1,y_2\}$, $\Y_{s_2} = \{y_3,y_4,y_5\}$, $\Y_{s_3} = \{y_6\}$, and $\Y_{s_8} = \{y_{10}, y_{11}\}$, while the remaining elements are the singletons $y_6$, $y_7$, $y_8$, and $y_9$. Clusters coming from different actions cannot be merged together because of different labeling of the hidden state, where, e.g., $x_2$ may be labeled differently depending on whether action $a_1$ or $a_2$ is used.
}
\label{fig:observation.policy}
\end{figure}

A rich-observation MDP (\richmdp) (Fig.~\ref{fig:Gmodel}) is a tuple $M=\langle\mathcal{X},\mathcal{Y},\mathcal{A},R, f_T,f_O\rangle$, where $\mathcal{X}$, $\mathcal{Y}$, and $\mathcal{A}$ are the sets of hidden states, observations, and actions. We denote by $X$, $Y$, $A$ their cardinality and we enumerate their elements by $i\in[X]=\{1..X\}$, $j\in[Y]=\{1..Y\}$, $l\in[A]=\{1..A\}$. We assume that the hidden states are fewer than the observations, i.e., $X \leq Y$. We consider rewards bounded in $[0,1]$ that depend only on hidden states and actions with a reward matrix $R\in \mathbb{R}^{A\times X}$ such that $[R]_{i,l}=\mathbb{E}[r(x=i,a=l)]$. The dynamics of the MDP is defined on the hidden states as $T_{i',i,l}:=f_T(i'|i,l)=\Prob(x'\!\!=\!i'|x\!=\!i,a\!=\!l)$, where $T\in\mathbb{R}^{X\times X\times A}$ is the transition tensor. The observations are generated as $[O]_{j,i}=f_O(j|i)=\Prob(y\!=\!j|x\!=\!i)$, where the observation matrix $O\in \mathbb{R}^{Y\times X}$ has minimum \textit{non-zero} entry $O_{\min}$.  
This model is a strict subset of POMDPs since each observation $y$ can be generated by only one hidden state (see Fig.~\ref{fig:observation.policy}-\textit{left}) and thus $\X$ can be seen as a non-overlapping clustering of the observations. 

We denote by $\mathcal{Y}_x = \mathcal{Y}_i = \{y=j\in\mathcal{Y}: [O]_{j,i} > 0\}$ the set of observations in cluster $x$, while $x_y = x_j$ is the cluster observation $y=j$ belongs to.\footnote{Throughout the paper we use the indices $i$, $j$, and $l$ and the ``symbolic'' values $x$, $y$, and $a$ interchangeably.} 
This structure implies the existence of an \textit{observable} MDP $M_{\Y} = \langle \mathcal {Y}, \mathcal{A}, R', f_T'\rangle$, where $R' = R$ as the reward of an observation-action pair $(y,a)$ is the same as in the hidden state-action pair $(x_y,a)$, and the dynamics can be obtained as $f_T'(j'|j,a) = \Prob(y'\!\!=\!j' | y\!=\!j,a\!=\!l) = \Prob(y'\!\!=\!j' | x'\!\!=\!x_{j'}) \Prob(x'\!\!=\!x_{j'} | x\!=\!x_j, a=l) = [O]_{j',x_{j'}} [T]_{x_{j'},x_j,l}$. We measure the performance of an observation-based policy $\pi_{\Y}:\Y\rightarrow \A$ starting from a hidden state $x$ by its asymptotic average reward $\rho(x; \pi_\Y) = \lim_{N\rightarrow \infty} \mathbb{E}\big[\sum_{t=1}^N r_t /N \big| x_0 = x, \pi_{\Y}\big]$.
Given the mapping between the \richmdp to the hidden MDP, the optimal policy $\pi_{\Y}^*(y)$ is equal to the optimal hidden-state policy $\pi^*_{\X}: \X\rightarrow \A$ for all $y\in\Y_x$. 
The learning performance of an algorithm run over $N$ steps is measured by the regret
\begin{align*}
R_N = N\rho^* - \big[\sum_{t=1}^N r_t\big],~ \textit{where } \rho^* = \rho(\pi_{\X}^*).
\end{align*}
Finally we recall that the diameter of the observation MDP is defined as 
\begin{align*}
D_{\Y} = \max_{y,y'\in\Y} \min_{\pi:\Y\rightarrow \A} \mathbb{E}\big[\tau_{\pi}(y,y')\big],
\end{align*}
where $\tau_{\pi}(y,y')$ is the (random) number of steps from $y$ to $y'$ by following the observation-based policy $\pi$ (similar for the diameter of the hidden MDP). 

%% file: 4RLalgorithm.tex

\section{Learning \richmdp}\label{sec:stochastic}


In this section we introduce the spectral method used to learn the structure of the observation matrix $O$. In particular, we show that we do not need to estimate $O$ exactly as the clusters $\{\Y_x\}_{x\in\X}$ can be recovered by identifying the non-zero entries of $O$.
%
%
%
We need a first assumption on the \richmdp.

\begin{assumption}\label{asm:ergodicity}
The Markov chain induced on the hidden MDP $M$ by any policy $\pi_{\Y}$ is ergodic.
\end{assumption}

Under this assumption for any policy $\pi$ there is a stationary distribution over hidden states $\omega_{\pi}$ and a stationary distribution conditional on an action $\omega_{\pi}^{(l)}(i) = \Prob_\pi(x\!=\!i | a\!=\!l)$. 
 Let $\X_{\pi}^{(l)} = \{i\in[X]: \omega_{\pi}^{(l)}(i) > 0\}$ be the hidden states where action $l$ could be taken according to policy $\pi$. In other words, if $\Y_{\pi}^{(l)} = \{j\in[Y]: \pi(j)=l\}$ is the set of observations in which policy $\pi$ takes action $l$, then $\X_{\pi}^{(l)}$ is the set of hidden states $\{x_y\}$ with $y\in\Y_{\pi}^{(l)}$ (see Fig.~\ref{fig:observation.policy}-\textit{right}). We also define the set of all hidden states that can be reached starting from states in $\X_{\pi}^{(l)}$ and taking action $l$, that is
 \setlength{\abovedisplayskip}{5pt}
\setlength{\belowdisplayskip}{5pt}
\begin{align*} 
\wb\X_{\pi}^{(l)} = \bigcup_{i\in\X_{\pi}^{(l)}}\!\!\! \Big\{i'\in[X]: \Prob\big(x'=i' | x=i, a=l\big)>0\Big\}.
\end{align*}
Similarly $\underline\X_{\pi}^{(l)}$ is the set of hidden states from which we can achieve the states $\X_{\pi}^{(l)}$ by policy $\pi$.
We need the following assumption.

\begin{assumption}[Full-Rank]\label{asm:expansive}
Given any action $l$, the slice of transition tensor $[T]_{\cdot,\cdot,l}$ is full rank.
\end{assumption}

Asm.~\ref{asm:expansive} implies that for any action $l$ the dynamics of $M$ is ``expansive'', i.e., $|\X_{\pi}^{(l)}| \leq |\wb\X_{\pi}^{(l)}|$.
In other words, the number of hidden states where policy $\pi$ can take an action $l$ (i.e., $\X_{\pi}^{(l)}$) is smaller than the number of states that can be reached when executing action $l$ itself (i.e., $\wb\X_{\pi}^{(l)}$). These two assumptions ensure that the underlying Markov process is stochastic.

\textbf{Multi-view model and exact recovery.}
We are now ready to introduce the multi-view model~\citep{anandkumar2014tensor} that allows us to reconstruct the clustering structure of the \richmdp Alg.~\ref{alg:spectral.learning}. We consider the trajectory of observations and actions generated by an arbitrary policy $\pi$ and we focus on three consecutive observations $y_{t-1}, y_t, y_{t+1}$ at any step $t$. As customary in multi-view models, we \textit{vectorize} the observations into three one-hot view vectors $\vec{v}_1$, $\vec{v}_2$, $\vec{v}_3$ in $\{0,1\}^Y$ such that $\vec{v}_1 = \vec{e}_j$ means that the observation in the first view is $j\in[Y]$ and where we remap time indices $t-1, t, t+1$ onto $1$, $2$, and $3$. We notice that these views are indeed independent random variables when conditioning on the state $x_2$ (i.e., the hidden state at time $t$) and the action $a_2$ (i.e., the action at time $t$), thus defining a multi-view model for the hidden state process. Let $k_1 = |\underline\X_{\pi}^{(l)}|$, $k_2 = |\X_{\pi}^{(l)}|$ and $k_3 = |\wb \X_{\pi}^{(l)}|$, then we define the factor matrices $V_1^{(l)}\!\in\mathbb{R}^{Y\times k_1},V_2^{(l)}\!\in\mathbb{R}^{Y\times k_2},V_3^{(l)}\!\in\mathbb{R}^{Y\times k_3}$ as follows
\begin{align*}
[V_p^{(l)}]_{j,i}=\Prob(\vecv_p=\vec{e}_j|x_2=i,a_2=l), 
\end{align*}
where for $p\!=\!1,~i\in\underline\X_{\pi}^{(l)}$, for $p\!=\!2,~i\in\X_{\pi}^{(l)}$, and for $p\!=\!3,~ i\in\wb\X_{\pi}^{(l)}.$
%

We are interested in estimating $V_2^{(l)}$ since it directly relates to the observation matrix as
\begin{align}\label{eq:v2toO}
[V_2^{(l)}]_{j,i} = \frac{\Prob(a_2=l | y_2=j) \Prob(y_2=j | x_2=i)}{\Prob(a_2=l | x_2=i)}
= \frac{\mathbb{I}\{\pi(j)=l\} f_O(j|i)}{\Prob(a_2=l | x_2=i)},
\end{align}
where $\mathbb{I}$ is the indicator function. As it can be noticed, $V_2^{(l)}$ borrows the same structure as the observation matrix $O$ and since we want to recover only the clustering structure of $M$ (i.e., $\{\Y_i\}_{i\in[X]}$), it is sufficient to compute the columns of $V_2^{(l)}$ up to any multiplicative constant. In fact, any non-zero entry of $V_2^{(l)}$ corresponds to a non-zero element in the original observation matrix (i.e., $[V_2^{(l)}]_{j,i} > 0 \Rightarrow [O]_{j,i} > 0$) and for any hidden state $i$, we can construct a cluster $\Y_i^{(l)} = \{j\in[Y]: [V_2^{(l)}]_{j,i} > 0\}$, which is accurate up to a re-labelling of the states. More formally, there exists a mapping function $\sigma^{(l)}:\X\rightarrow \X$ such that any pair of observations $j,j'\in\Y_i^{(l)}$ is such that $j,j'\in\Y_{\sigma(i)}$. Nonetheless, as illustrated in Fig.~\ref{fig:observation.policy}-\textit{right}, the clustering may not be minimal. In fact, we have $[O]_{j,i} > 0 \not\Rightarrow [V_2^{(l)}]_{j,i} > 0$ since $[V_2^{(l)}]_{j,i}$ may be zero because of policy $\pi$, even if $[O]_{j,i} > 0$. Since the (unknown) mapping function $\sigma^{(l)}$ changes with actions, we are unable to correctly ``align'' the clusters and we may obtain more clusters than hidden states. We define $\S$ as the auxiliary state space obtained by the partial aggregation and we prove the following result.
%

\begin{figure}
\vspace*{-0.1in}
\begin{algorithm}[H]
\begin{small}
\setstretch{1.3}
\begin{algorithmic}
\STATE \hspace{-0.1in}\textbf{Input:} Trajectory $(y_1,a_1,\ldots,y_N)$
\STATE \hspace{-0.1in}\textbf{For} Action $l\in[A]$ \textbf{do}
\STATE \hspace{-0.05in}Estimate second moments $\wh{K}_{2,3}^{(l)}$, $\wh{K}_{1,3}^{(l)}$, $\wh{K}_{2,1}^{(l)}$, and $\wh{K}_{3,1}^{(l)}$
\STATE \hspace{-0.05in}Estimate the rank of matrix $\wh{K}_{2,3}^{(l)}$ (see App.~\ref{sec:rank})
\STATE \hspace{-0.05in}Compute symmetrized views $\wt v_{1,t}$ and $\wt v_{3,t}$, for $t=2..N-2$
\STATE \hspace{-0.05in}Compute second and third moments $\wh M_{2}^{(l)}$ and $\wh M_{3}^{(l)}$
\STATE \hspace{-0.05in}Compute $\wh V_2^{(l)}$ from the tensor decomposition of (an orthogonalized version of) $\wh M_{3}^{(l)}$
\STATE \hspace{-0.05in} return clusters
$$\wh{\Y}_i^{(l)} = \{j\in[Y]: [\wt V_2^{(l)}]_{j,i} > 0\}$$
\end{algorithmic}
\caption{Spectral learning algorithm.}
\label{alg:spectral.learning}
\end{small}
\end{algorithm}
\vspace*{-0.5cm}
\end{figure}

\begin{lemma}\label{lem:partial.cluster}
Given a policy $\pi$, for any action $l$ and any hidden state $i\in\X_{\pi}^{(l)}$, let $\Y_i^{(l)}$ be the observations that can be clustered together according to $V_2^{(l)}$ and $\Y^{\mathsf{c}} = \Y \setminus \bigcup_{i,l}\Y_i^{(l)}$ be the observations not clustered, then the auxiliary state space $\S$ contains all the clusters $\{\bigcup_{i,l}\Y_i^{(l)}\}$ and the singletons in $\Y^{\mathsf{c}}$ for a total number of elements $S = |\S| \leq A X$.
\end{lemma}
%
\vspace*{-.1in}
We now show how to recover the factor matrix $V_2^{(l)}$. We introduce mixed second and third order moments as
$K^{(l)}_{p,q}=\mathbb{E}[\vecv_p\otimes \vecv_q], K^{(l)}_{p,q,r}=\mathbb{E}[\vecv_p\otimes \vecv_q\otimes\vecv_r]$
where $p,q,r$ is any permutation of $\lbrace 1,2,3\rbrace$. Exploiting the conditional independence of the views, the second moments can be written as
%
\begin{align*}
K^{(l)}_{p,q} = \sum_{i\in\X^{l}_{\pi}} \omega_\pi^{(l)}(i) [V_p^{(l)}]_{:,i} \otimes [V_q^{(l)}]_{:,i}
\end{align*}
%
where $[V_p^{(l)}]_{:,i}$ is the $i$-th column of $V_p^{(l)}$. 
In general the second moment matrices are rank deficient, with rank $X_{\pi}^{(l)}$. We can construct a symmetric second moment by introducing the symmetrized views 
\begin{align}\label{eq:symmetric.views}
\wt v_1 = K^{(l)}_{2,3} (K^{(l)}_{1,3})^{\dagger}\vecv_1,\enspace\enspace\quad \wt v_3 = K^{(l)}_{2,1}(K^{(l)}_{3,1})^{\dagger}\vecv_3,
\end{align}
where $K^\dagger$ denotes the pseudoinverse. Then we can construct the second and third moments as 
\begin{align}\label{eq:second.moment}
M_2^{(l)} \!=\! \E\big[\wt v_1 \otimes \wt v_3 \big] \!=\! \sum_{i\in\X^{(l)}_{\pi}} \!\!\omega_\pi^{(l)}(i) [V_2^{(l)}]_{:,i} \otimes [V_2^{(l)}]_{:,i}.
\end{align}
%
%
\begin{align}\label{eq:third.moment}
M_3^{(l)} = \E\big[\wt v_1 \otimes \wt v_3 \otimes \vecv_2 \big]
= \sum_{i\in\X^{l}_{\pi}}\omega_\pi^{(l)}(i) [V_2^{(l)}]_{:,i} \otimes [V_2^{(l)}]_{:,i}\otimes [V_2^{(l)}]_{:,i}.
\end{align}
We can now employ the standard machinery of tensor decomposition methods to orthogonalize the tensor $M_3^{(l)}$ using $M_2^{(l)}$ and recover $V_2^{(l)}$ (refer to~\citep{anandkumar2014tensor} for further details) and a suitable clustering. 
\begin{lemma}\label{lem:tensor.decomposition}
For any action $l\in[A]$, let $M_3^{(l)}$ be the third moment constructed on the symmetrized views as in Eq.~\ref{eq:third.moment}, then we can orthogonalize it using the second moment $M_2^{(l)}$ and obtain a unique spectral decomposition from which we compute the exact factor matrix $[V_2^{(l)}]_{j,i}$. As a result, for any hidden state $i\in\X_\pi^{(l)}$ we define the cluster $\wt{\Y}_i^{(l)}$ as
\begin{align}\label{eq:estimated.cluster2}
\wt{\Y}_i^{(l)} = \{j\in[Y]: [V_2^{(l)}]_{j,i} > 0\}
\end{align}
and there exists a mapping $\sigma^{(l)}:X\rightarrow X$ such that if $j,j'\in \wt{\Y}_i^{(l)}$ then $j,j'\in \Y_{\sigma^{(l)}(i)}$ (i.e., observations that are clustered together in $\wt{\Y}_i^{(l)}$ are clustered in the original \richmdp).
\end{lemma}


%
\begin{figure}
\vspace*{-0.5cm}
\hspace*{.2in}
\begin{algorithm}[H]
\begin{small}
\setstretch{1.2}
\begin{algorithmic}
\STATE \hspace{-0.1in}\textbf{Initialize:} $t=1$, initial state $x_1$, $k=1$, $\delta/N^6$
\STATE \hspace{-0.1in}\textbf{While} {$t < N$} \textbf{do}
\STATE \hspace{-0.05in}Run Alg.~\ref{alg:spectral.learning} on samples from epoch $k-1$ and obtain $\wh\S$
\STATE \hspace{-0.05in}Compute aux. space ${\wh\S}^{(k)}$ by merging $\wh\S$ and $\wh\S^{(k-1)}$
\STATE \hspace{-0.05in}Compute the estimate reward $r^{(k)}$ and dynamics $p^{(k)}$
\STATE \hspace{-0.05in}Construct admissible AuxMDPs $\mathcal{M}^{(k)}$ 
\STATE \hspace{-0.05in}Compute the optimistic policy 
\begin{align}\label{eq:optimism}
\wt{\pi}^{(k)} = \arg\max\limits_{\pi}\max\limits_{M\in\mathcal{M}^{(k)}} \rho(\pi; M)
\end{align}  
\STATE \hspace{-0.05in}Set $v^{(k)}( s,l) = 0$ for all actions $l\in\mathcal{A}, s\in {\wh\S}^{(k)}$
\STATE \hspace{-0.05in}\textbf{While} $\forall l, \forall  s, v^{(k)}(s,l) < \max\lbrace 1,N^{(k)}(s,l)\rbrace$ \textbf{do}
\STATE Execute $a_t = \wt\pi^{(k)}(s_t)$
\STATE Observe reward $r_t$ and observation $y_t$
\end{algorithmic}
\caption{Spectral-Learning UCRL(\scucrl).}
\label{alg:sm.ucrl}
\end{small}
\end{algorithm}
\vspace{-0.3in}
\end{figure}

The computation complexity of Alg.~\ref{alg:spectral.learning} has been studied by~\citet{song2013nonparametric} and is polynomial in the rank of third order moment.

\textbf{Spectral learning.} \footnote{We report the spectral learning algorithm for the tensor decomposition but a very similar algorithm and guarantees can be derived for the matrix decomposition approach when the eigenvalues of $\wh M_2^{(l)}$ for all actions and all posible policy have multiplicity 1. This further condition is not required when the tensor decomposition is deployed.} 
While in practice we do not have the exact moments, we can only estimates them through samples. Let $N$ be the length of the trajectory generated by policy $\pi$, then we can construct $N-2$ triples $\{y_{t-1}, y_t, y_{t+1}\}$ that can be used to construct the corresponding views $\vecv_{1,t},\vecv_{2,t},\vecv_{3,t}$ and to estimate second mixed moments as
\begin{align*}
\wh{K}^{(l)}_{p,q}=\frac{1}{N{(l)}} \sum_{t=1}^{N(l)-1}\mathbb{I}(a_t=l) ~\vecv_{p,t} \otimes \vecv_{q,t},
\end{align*}
with $p,q\in\lbrace 1,2,3\rbrace$ and $N{(l)}=\sum_t^{N-1}\mathbb{I}(a_t=l)$. Furthermore, we require knowing $|\X_{\pi}^{(l)}|$, which is not known apriori. Under Asm.~\ref{asm:ergodicity} and~\ref{asm:expansive}, for any action $l$, the rank of ${K}^{(l)}_{2,3}$ is indeed $|\X_{\pi}^{(l)}|$ and thus $\wh{K}^{(l)}_{2,3}$ can be used to recover the rank. The actual way to calculate the efficient rank of $\wh{K}^{(l)}_{2,3}$ is quite intricate and we postpone the details to App.~\ref{sec:rank}.
From $\wh{K}^{(l)}_{p,q}$ we can construct the symmetric views $\wt v_{1,t}$ and $\wt v_{3,t}$ as in Eq.~\ref{eq:symmetric.views} and compute the estimates of second and third moments as

\begin{align*}
\wh{M}_2^{(l)} &= \frac{1}{N{(l)}}  \sum_{t=1}^{N-1} \mathbb{I}(a_t=l)\wt v_{1,t} \otimes \wt v_{3,t},\\
\wh{M}_3^{(l)} &= \frac{1}{N{(l)}}  \sum_{t=1}^{N-1} \mathbb{I}(a_t=l)\wt v_{1,t} \otimes \wt v_{3,t} \otimes \vecv_{2,t}.
\end{align*}

Following the same procedure as in the exact case, we are then able to recover estimates of the factor matrix $\wh V_2^{(l)}$, which enjoys the following error guarantee.

\begin{lemma}\label{lem:decomp}
Under Asm.~\ref{asm:ergodicity} and~\ref{asm:expansive}, let $\widehat{V}^{(l)}_2$ be the empirical estimate of ${V}^{(l)}_2$ obtained using $N$ samples generated by a policy $\pi$. There exists $N_0$ such that for any $N(l) > N_0$, $l\in\A$, $i\in \X_{\pi}^{(l)}$ w.p. $1-\delta$
\begin{align}\label{eq:BO}
\|[V_2^{(l)}]_{\cdot,i}\!-\![\wh{V}_2^{(l)}]_{\cdot,i}\|_2\!\leq\! C_2\sqrt{\frac{\log(2Y^{3/2}/\delta)}{N(l)}}:=\BO^{(l)}
\end{align}
where $C_2$ is a problem-dependent constant independent from the number of observations $Y$.
\end{lemma}

While this estimate could be directly used to construct a clustering of observations, the noise in the empirical estimates might lead to $[\wh V_2^{(l)}]_{j,i} > 0$ for any $(j,i)$ pair, which prevents us from generating any meaningful clustering. On the other hand, we can use the guarantee in Lem.~\ref{lem:decomp} to single-out the entries of $\wh V_2^{(l)}$ that are non-zero w.h.p. We define the binary matrix $\wt V_2^{(l)}\in\{0,1\}^{Y\times X}$ as
\begin{align*}
[\wt V_2^{(l)}]_{j,i} = \begin{cases} 1 &\mbox{if } [\wh V_2^{(l)}]_{j,i} \geq \BO^{(l)} \\ 
0 & \mbox{otherwise} \end{cases},
\end{align*}
which relies on the fact that $[\wh V_2^{(l)}]_{j,i} -\BO^{(l)} > 0$ implies $[V_2^{(l)}]_{j,i} > 0$. At this point, for any $l$ and any $i\in\X_\pi^{(l)}$, we can generate the cluster 
\begin{align}\label{eq:estimated.cluster3}
\wh{\Y}_i^{(l)} = \{j\in[Y]: [\wt V_2^{(l)}]_{j,i} > 0\},
\end{align}
which is guaranteed to aggregate observations correctly in high-probability. We denote be $\wh\Y^{\mathsf{c}} = \Y \setminus \bigcup_{i,l}\wh{\Y}_i^{(l)}$ the set of observations which are not clustered through this process. Then we define the auxiliary state space $\wh\S$ obtained by enumerating all the elements of non-clustered observations together with clusters $\{\wh{\Y}_i^{(l)}\}_{i,l}$, for which we have the following guarantee.

\begin{corollary}\label{cor:partial.cluster.est}
Let $\wh\S$ be the auxiliary states composed of clusters $\{\wh{\Y}_i^{(l)}\}$ and singletons in $\Y^{\mathsf{c}}$ obtained by clustering observations according to $\wt V_2^{(l)}$, then for any pair of observations $j,j'$ clustered together in $\wh \S$, there exists a hidden state $i$ such that $j,j'\in \Y_i$. Finally, $\wh\S \rightarrow \S$ as $N$ tends to infinity.
\end{corollary}
\section{RL in \richmdp}
\begin{figure}[t]
\centering
\includegraphics[height=4.cm]{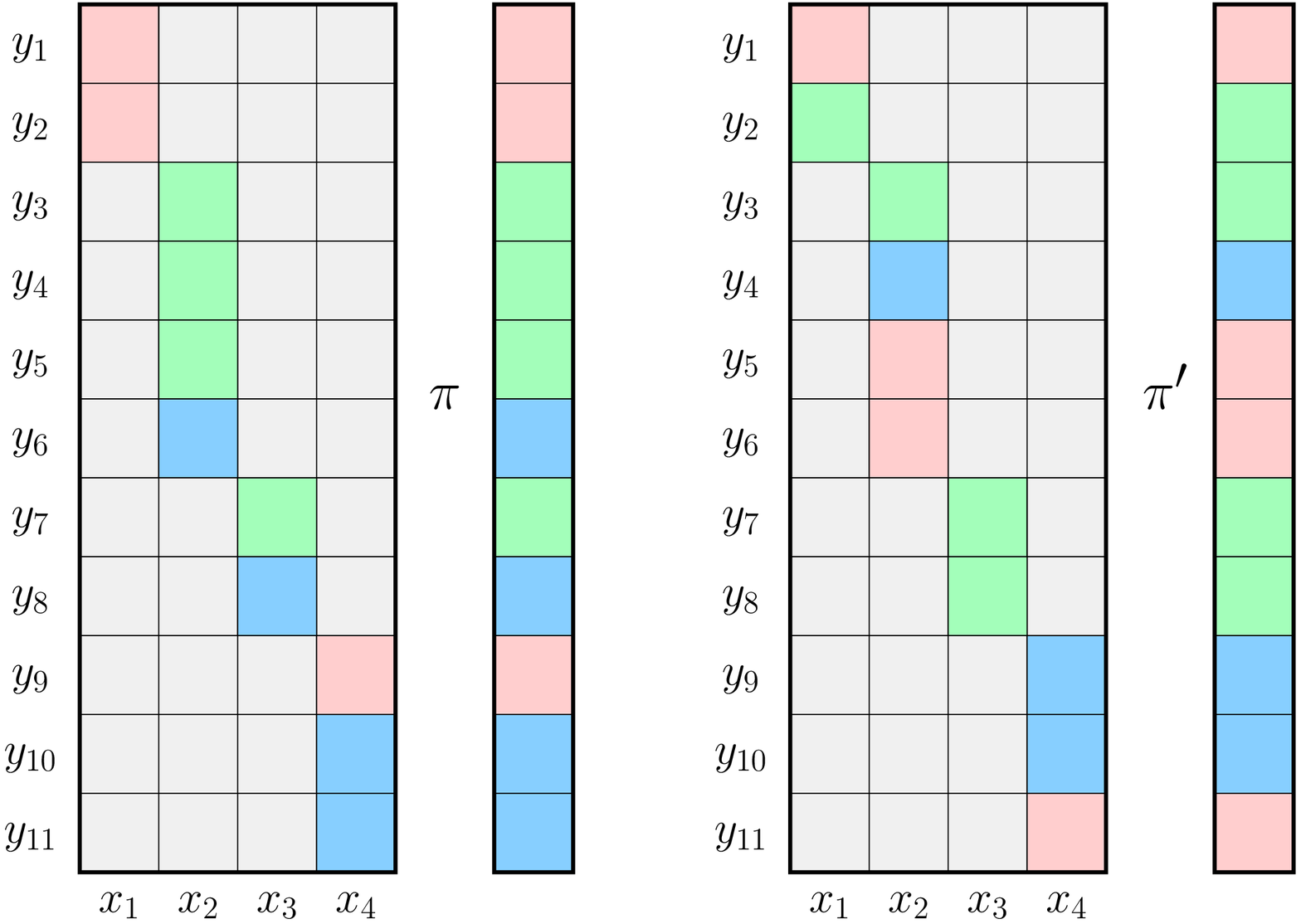}
\caption{ Examples of clusterings obtained from two policies that can be effectively merged.}
\label{fig:mappolicy}
\vspace{-.0in}
\end{figure}

 We now describe the spectral learning UCRL (\scucrl) (Alg.~\ref{alg:sm.ucrl}) obtained by integrating the spectral method above with the UCRL strategy. 
The learning process is split into epochs of increasing length. At the beginning of epoch $k$, we use the trajectory $(s_1,a_1,..,s_{N^{(k-1)}})$ generated at previous epoch using auxiliary states $s\in\wh\S^{(k)}$ to construct the auxiliary state space ${\wh\S}$ using Alg.~\ref{alg:spectral.learning}.\footnote{Since Alg.~\ref{alg:spectral.learning} receives as input a sequence of auxiliary states rather than observations as in Sect.~\ref{sec:stochastic} the spectral decomposition runs on a space of size $|\wh S^{(k-1)}|$ instead of $Y$, thus reducing the computation complexity.} As discussed in the previous section, the limited number of samples and the specific policy executed at epoch $k-1$ may prevent from clustering many observations together, which means that despite ${\wh\S}$ being a \textit{correct} clustering (see Cor.~\ref{cor:partial.cluster.est}), its size may still be large. While clusterings obtained at different epochs cannot be ``aligned'' because of different labelling, we can still effectively merge together any two clusterings $\wh\S$ and $\wh\S'$ generated by two different policies $\pi$ and $\pi'$. 
We illustrate this procedure through Fig.~\ref{fig:mappolicy}. Observations $y_3$, $y_4$, and $y_5$ are clustered together in the auxiliary space generated by $\pi$, while $y_5$ and $y_6$ are clustered together using $\pi'$. While the labeling of the auxiliary states is arbitrary, observations preserve their labels across epochs and thus we can safely conclude that observations $y_3$, $y_4$, $y_5$, and $y_6$ belong to the same hidden state. Similarly, we can construct a new cluster with $y_9$, $y_{10}$, and $y_{11}$, which, in this case, returns the exact hidden space $\X$. Following this procedure we generate $\wh\S^{(k)}$ as the clustering obtain by merging $\wh\S$ and $\wh\S^{(k-1)}$ (where $\wh\S^{1} = \Y$). 

\begin{figure}[ht]
\centering
\includegraphics[height=4.0cm]{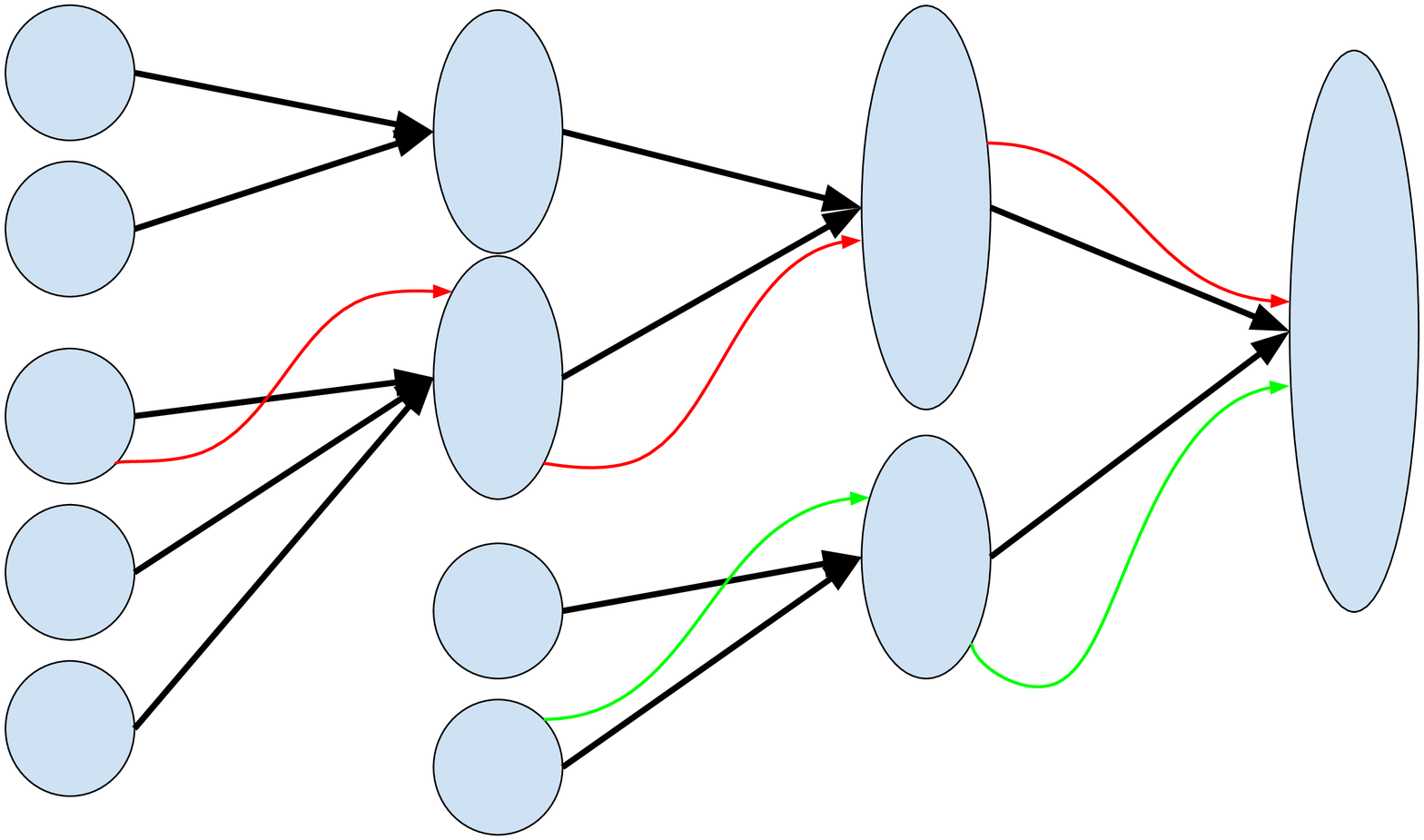}
\caption{Monotonic evolution of clusters, each layer is the beginning of an epoch. The green and red paths are two examples for two different cluster aggregation.}
\label{fig:evolution}
\end{figure}

At this point we can directly estimate the reward and transition model of the auxiliary MDP constructed on $\wh\S^{(k)}$ by using empirical estimators. For a sequence of clustering $\wh\S^{(0)},\ldots, \wh\S^{(k)}$, since the clustering $\wh\S^{(k)}$ is \textit{monotonic} (i.e., observations clustered at epoch $k$ stay clustered at any other epoch $k' > k$) any cluster $s^{(t^k)} \in \wh\S^{(k)}$ can be represented as result of a \textit{monotonically} aggregating observations as a increasing series of $s^1\subseteq s^2\subseteq \ldots \subseteq s^{(t^k-1)}$ (not unique, and random. As it is has been shown in Fig.~\ref{fig:evolution} any branching can be considered as one of these series. Let's choose one of them.
Here, $s^t$ is a cluster at time point $t(\leq t_k)$ which evolves to the cluster $s^{(t_k)}$. For a clustering sequence $s^1\subseteq s^2\subseteq \ldots \subseteq s^{(t^k-1)}$, evolving to $s^{(t^k)}$, define $N^{(k)}(s,a)$, the number of samples in interest is:
\begin{align*}
N^{(k)}(s^{(t^k)},a) = \sum_t^{t^{(k)}}\mathbbm{1}(y_t\in s^t)\mathbbm{1}(a_t = a)
\end{align*}
with an abuse of notation, we write $y\in s^t$ to denote that the observation $y$ has been clustered into an auxiliary state $s^t$ at time step $t$. For any observation $y$, we use all the samples of $y$ to decide whether to merge this observation to any cluster. When we merge this observation to a cluster, we do not use the past sample of $y$ for the empirical estimates of reward and transition. For example, Fig.~\ref{fig:mappolicy}, we cluster together $\lbrace y_3, y_4,y_5\rbrace$. At the beginning of each epoch, we use all the samples to decide whether $y_6$ belongs to this cluster. For an epoch, when we decide that $y_6$ belongs to this cluster, we do not use the samples of $y_6$ up to this epoch to estimate the reward and transition estimates of cluster $\lbrace y_3, y_4,y_5,y_6\rbrace$.
Therefore, to estimate the empirical mean of reward and the transition kernel, we have
\begin{align*}
\wh r^{(k)}(s^{(t^k)},a) = \sum_t^{t^{(k)}}r_t \mathbbm{1}(y_t\in s^t)\mathbbm{1}(a_t = a)/N^{(k)}(s,a)
\end{align*}
and for transitions, let's define the following count
\begin{align*}
N^{(k)}(s^{(t^k)},a,s') = \sum_t^{t^{(k)}}\mathbbm{1}(y_{t+1}\in s')\mathbbm{1}(y_t\in s^t)\mathbbm{1}(a_t = a)
\end{align*}
therefore
\begin{align*}
\wh p^{(k)}(s' | s^{(t^k)},a) = N^{(k)}(s^{(t^k)},a,s')/N^{(k)}(s^{(t^k)},a)
\end{align*}
 then we return the estimates.\footnote{Since the clustering $\wh\S^{(k)}$ is \textit{monotonic}, $\wh r^{(k)}$ and $\wh p^{(k)}$ can be computed incrementally without storing the statistics $N^{(k)}(y,a,y')$, $N^{(k)}(y,a)$, and $R^{(k)}(y,a)$ at observation level, thus significantly reducing the space complexity of the algorithm.}
For further use, we define the per-epoch samples of interest for  $s\in\wh\S^{(k)}$ as $\nu^{(k)}( s^{(t^k)},a) := \sum_{y\in\Y}\sum_{t^{(k-1)}}^{t^{(k)}}z_{t,t^{(k)}}(y)$
%
%
The corresponding confidence intervals are such that for any $s\in {\wh\S}^{(k)}$ and $a\in\mathcal{A}$
%
%
\begin{align*}
&\|p(\cdot|s,a)\text{--}\widehat{p}^{(k)}(\cdot|s,a)\|_1 \!\leq\! d_p(s,a)\!=\! \sqrt{\frac{28S^{(k)}\log(\frac{2AN^{(k)}}{\delta})}{N^{(k)}(s,a)}},\\
&|\bar{r}(s,a)-\widehat{r}^{(k)}(s,a)|\leq d_r(s,a)= \sqrt{\frac{28\log(\frac{2YAN^{(k)}}{\delta})}{N^{(k)}(s,a)}},
\end{align*} 

hold w.p. $1-\delta$, where $p(\cdot|s,a)$ and $\bar{r}$ are the transition probabilities and reward of the auxiliary MDP $M_{\wh\S^{(k)}}$ App.~\ref{Apx:concentration}. Given the estimates and the confidence intervals, we construct a set of plausible auxiliary MDPs, $\M^{(k)}$, where the reward means and transition probabilities satisfy the confidence intervals. 

At this point we can simply apply the same steps as in standard UCRL, where an optimistic auxiliary MDP $\wt{M}^{(k)}$ is constructed using the confidence intervals above and extended value iteration (EVI)~\citep{jaksch2010near}. The resulting optimal optimistic policy $\wt\pi^{(k)}$ is then executed until the number of samples at least for one pair of auxiliary state and action is doubled. 

EVI has a per-iteration complexity which scales as $\mathcal{O}((\wh S^{(k)})^2A)$ thus gradually reducing the complexity of \ucrl on the observation space (i.e., $\mathcal{O}((Y)^2A)$) as soon as observations are clustered together. When the whole clustering is learnt, the computational complexity of EVI tends to $\mathcal{O}((X)^2A)$. Moreover, since we aggregate the samples of the elements in clusters, therefore more accurate estimates, the number of times we call EVI algorithm goes from $\mathcal{O}(Y\log(N))$ to $\mathcal{O}(X\log(N))$.

\begin{theorem}\label{thm:regret}
Consider a \richmdp $M=\langle\mathcal{X},\mathcal{Y},\mathcal{A},R,f_T,f_O\rangle$ with diameter $D_{\X}$. If \smcmdp is run over $N$ time steps, under Asm.~\ref{asm:ergodicity} and \ref{asm:expansive}, with probability $1-\delta$ it suffers the total regret of 
\begin{align*}
\textit{Reg}_{N}\leq \sum_{k= 1}^{K}\Big( D_{\wh S^{(k)}}\sqrt{\wh S^{(k)}\log\Big(\frac{N^{(k)}}{\delta}\Big)}\hspace*{-.4em}\sum_{s\in\mathcal{\wh S}^{(k)},a}\hspace*{-.4em}\frac{\nu^{(k)}( s,a)}{\sqrt{N^{(k)}(s,a)}}\Big),
\end{align*}
where $(\S^{(k)})$ is the sequence of auxiliary state spaces generated over $K$ epochs.
\end{theorem}

\textbf{Remark.} This bound shows that the per-step regret decreases over epochs. First we notice that only the regret over the first few (and short) epochs actually depends on the number of observations $Y$ and the diameter $D_\Y$. As soon as a few observations start being clustered into auxiliary states, the regret depends on the number of auxiliary states $\wh S^{(k)}$ and the diameter $D_{\S^{(k)}}$. Since $\wh S^{(k)}$ decreases every time an observation is added to a cluster and $D_{\S^{(k)}}$ is monotonically decreasing with of $\wh \S^{(k)}$, the per-step regret significantly decreases with epochs.\footnote{We refer to the per-step regret since an epoch may be longer, thus making the cumulative epoch regret larger.} Cor.~\ref{cor:partial.cluster.est} indeed guarantees that the number of auxiliary states in $\wh\S$ reduces down to $|\S|$ ($XA$ in the worst case) as epochs get longer. Furthermore we recall that even if the clustering $\wh \S$ returned by the spectral method is not minimal, merging clusters across epochs may rapidly result in very compact representations even after a few epochs.

\textbf{Minimal clustering.} While Thm.~\ref{thm:regret} shows that the performance of \smcmdp improves over epochs, it does not relate it to the (ideal) performance that could be achieved when the hidden space had been known. Unfortunately, even if the number of clusters in $\wh{\S}^{(k)}$ is nearly-minimal, the MDP constructed on the auxiliary state space may have a large diameter. In fact, it is enough that an observation $j$ with very low probability $O_{j,i}$ is not clustered (it is a singleton in $\S^{(k)}$) to have a diameter that scales as $1/O_{\min}$ (although its \textit{actual} impact on the regret may be negligible, for instance when $j$ is not visited by the current policy).
, in general the advantage obtained by clustering reduces the dependency on the number of states from $Y$ to $XA$ but it may not be effective in reducing the dependency on the diameter from $D_\Y$ to $D_\X$. 

In order to provide a minimal clustrting, we integrate Alg.~\ref{alg:sm.ucrl} with a clustering technique similar to the one used in~\cite{gentile2014online} and~\cite{ortner2013adaptive}. At any epoch $k$, we proceed by merging together all the auxiliary states in $\wh \S^{(k)}$ whose reward and transition confidence intervals overlap (i.e., $s$ and $s'$ are merged if the confidence interval $[\wh r(s,a) \pm d_r(s,a)]$ overlaps with $[\wh r(s',a) \pm d_r(s',a)]$ and $[\wh p(\cdot|s,a) \pm d_p(s,a)]$\footnote{Deviation $d_p(s,a)$ on a $\wh S$ dimensional simplex} overlaps with $[\wh p(\cdot|s',a) \pm d_p(s',a)]$. If the number of new clusters is equal to $X$, then we claim we learned the true clustering, if it is less than $X$ we ignore this temporary clustering and proceed to the next epoch. It is worth noting that this procedure requires the knowledge of $X$, while the spectral method, by its own, does not. While an explicit rate of clustering is very difficult to determine (the merging process depends on the spectral method, whose result depends on the policy, which in turn is determined according to the clustering at previous epochs), we derive worst-case bounds on the number of steps needed to start clustering at least one observation (i.e., steps before avoiding the dependency on $Y$ and $D_\Y$) and before the exact clustering is recovered.
\begin{corollary}\label{cor:regret}
Let $\tau_M=\max_{x,\pi}\mathbb{E}_{\pi}[\tau_\pi(x, x)]$ the maximum expected returning time in MDP $M$ (bounded due to ergodicity) and
\begin{align}\label{eq:firstlast}
\wb N_{\first} &= \frac{AY \tau_M}{O_{\min}}\frac{C_2\log(1/\delta)}{\max_{i,j}f_{O}(y=j|x=i)^2}; \nonumber\\
 \wb N_{\last} &= \frac{AY \tau_M}{O_{\min}^3}C_2\log(1/\delta).
\end{align}
After $\wb N_{\first}$ steps at least two observations are clustered and after $\wb N_{\last}$ steps all observations are clustered (but not necessarily in the minimum hidden space configuration) with probability $1-\delta$. This implies that after $\wb N_{\last}$ steps $|\wh{S}^{(k)}| \leq XA$. Furthermore, let $\gamma_r = \min_{x,x',a} |r(x,a)-r(x',a)|$ and $\gamma_p = \min_{x,x',a} \|p(\cdot|x,a) - p(\cdot|x',a)\|_1$ be the smallest gaps between rewards and transition probabilities and let $\gamma = \max\{\gamma_r, \gamma_p\}$ the maximum between the two.
In the worst case, using the additional clustering step together with \smcmdp guarantees that after $\wb N_{\X}$
\begin{align*}
 \min \Big\{\frac{AY^2 \tau_M}{\gamma^2}\log(1/\delta), \max\big\lbrace \frac{AS^2 \tau_M}{\gamma^2}\log(1/\delta) , \wb N_{\last}\big\rbrace\Big\}
\end{align*}
samples the hidden state $\X$ is correctly reconstructed (i.e., $\wh\S^{(k)} = \X$), therefore
\begin{align*}
\textit{Reg}_{N}&\leq 34D_{\X}X\sqrt{A(N-\wb N_{\X})\log(N/\delta)}\mathbb{I}(N\geq \wb N_{\X})\nonumber\\
&+\min\lbrace \wb N_{\X},34D_{\Y}Y\sqrt{A(\wb N_{\X})\log(N_{\X}/\delta)} \rbrace
\end{align*}
\end{corollary}
We first notice that this analysis is constructed over a series of worst-case steps (see proof in App.~\ref{app:clustering.rate}). Nonetheless, it first shows that the number of observations $Y$ does impact the regret only over the first $\wb N_{\first}$ steps, after which $\wh S^{(k)}$ is already smaller than $\Y$. Furthermore, after at most $\wb N_{\last}$ the auxiliary space has size at most $XA$ (while the diameter may still be as large as $D_{\Y}$). Finally, after $\wb N_{\X}$ steps $\wh{S}^{(k)}$ reduces to $\X$ and the performance of \smcmdp tends to the same performance of UCRL in the hidden MDP.

%% file: 4.1Experiment.tex
\section{Experiments}

\begin{figure}
\centering
\begin{psfrags}
\psfrag{sqrt(N)}[][1]{$\sqrt{N}$}
\psfrag{Regret}[][1]{Regret}
\psfrag{SL-UC}[][1]{\scalebox{.4}\scucrl}
\psfrag{DQN}[][1]{\scalebox{.4}{DQN}}
\psfrag{UCRL}[][1]{\scalebox{.4}{UCRL}}
\includegraphics[height=5.cm]{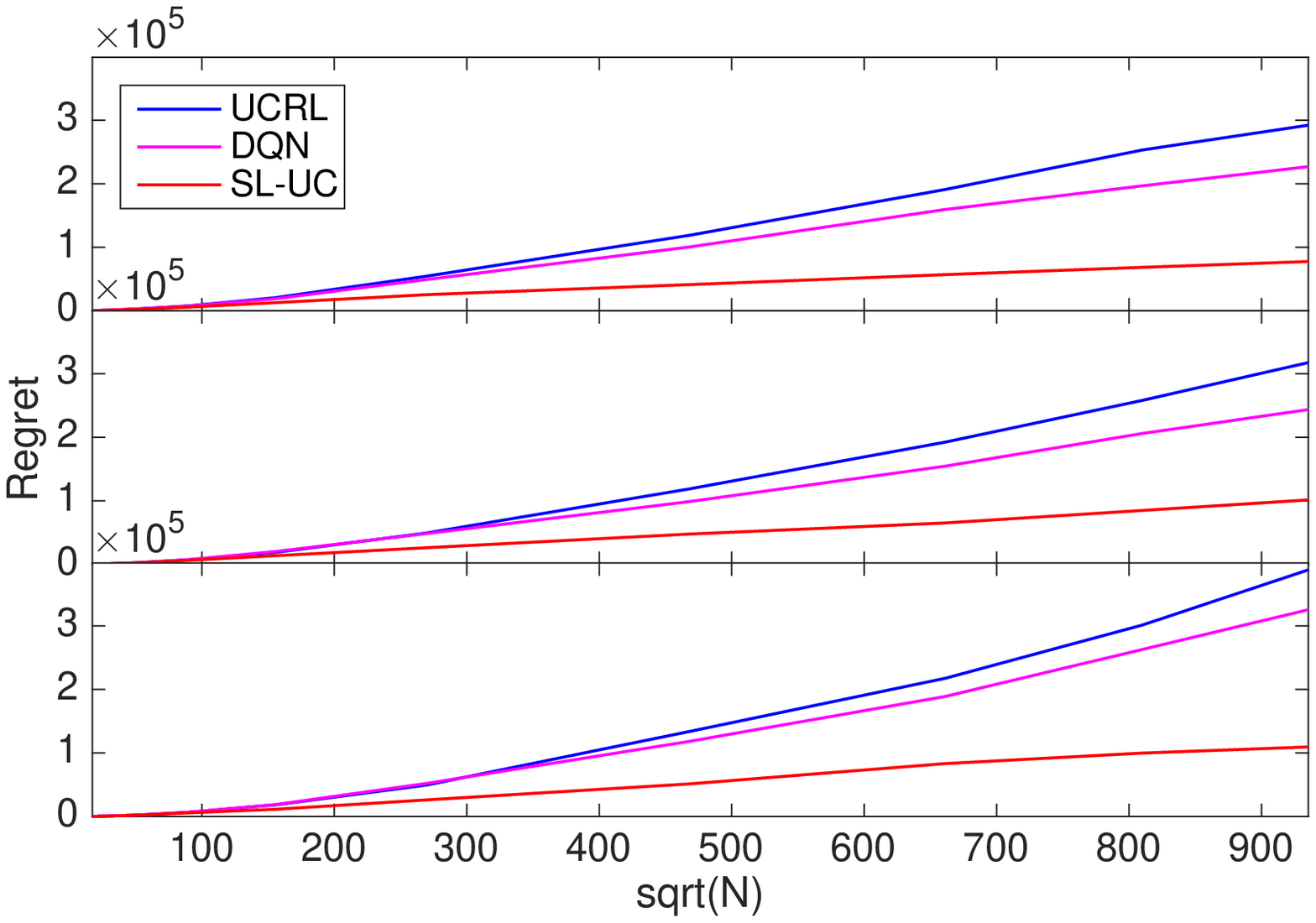}
 \end{psfrags}
\caption{Regret comparison for \richmdps with $X=5,A=4$ and from top to bottom $Y=10,20,30$.}
\label{fig:comparison}
\vspace{-.2in}
\end{figure}

We validate our theoretical results by comparing the performance of \smcmdp, UCRL2 (model based) and DQN (model free, function approximation)~\cite{mnih2013playing}, two well known RL algorithms. The goal of this experiment is to evaluate the dependency of regret to dimensionality of observation space. Generally, DQN is considered as model free RL method which extend the notion of Bellman residual \citep{antos2008learning} to deep RL. For DQN, we implement a three layers feed forward network (with no CNN block), equipped with RMSprop and replay buffer. We tune the hyper parameters of the network and report the best performance achieved by network of size $30\times 30\times 30$.

We consider three randomly generated \richmdps (Dirichlet transition and Uniform reward with different bias) with $X=5$, $A=4$ and observation spaces of sizes $Y=10, 20, 30$. Fig.~\ref{fig:comparison} reports the regret on a $\sqrt{N}$ scale where regret of UCRL and DQN grows much faster than \smcmdp's. While all regrets tend to be linear (i.e., growing as $\sqrt{N}$), we observe that the regret slope of UCRL and DQN are negatively affected by the increasing number of observations, while the regret slope of \smcmdp stays almost constant, confirming that the hidden space $\X$ is learned rapidly. These experiments are the first step towards more practical applications. Additional experiments App.~\ref{app:experiment}.

%% file: 5Conclusion.tex
\section{Conclusion}

We introduced \smcmdp, a novel RL algorithm to learn in \richmdps combining a spectral method for recovering the clustering structure of the problem and UCRL to effectively trade off exploration and exploitation. We proved theoretical guarantees showing that \smcmdp progressively refines the clustering so that its regret tends to the regret that could be achieved when the hidden structure is known in advance (in higher order term). Despite this result almost matching the regret obtained by running UCRL directly on the latent MDP, the regret analysis requires ergodicity of the MDP. 
One of the main open questions is whether the spectral clustering method could still provide ``useful'' clusterings when the state space is not fully visited (i.e., in case of non-ergodic MDP), so that observations are properly clustered where it is actually needed to learn the optimal policy App.~ \ref{Aux:deter}.

At the beginning of the learning, the algorithm deals with larger MDPs and gradually, while learning bigger cluster, it starts to deal with smaller MDPs. Reducing the state space of MDPs means lower cost in computing optimistic policy, having fewer number of epochs, and suffer from lower computation cost. Finally, this work opens several interesting
directions to extend the results for variety of state aggregation topologies \citep{li2006towards}. Furthermore, one
can aggregate the proposed method with other regret analyses, e.g. \citep{dann2017ubev}, \citep{azar2017minimax} to provide a new regret bound.

%% file: 6Supplementary.tex
\section{Proof of Lemma~\ref{lem:decomp}}

At the end of each epoch, e.g. $k$, we estimate the factor matrix $V^{(l)}_2$ (for all $l\in[A]$) using all the samples collected during that epoch according to policy $\wt\pi^{(k)}$. In order to simplify the notation, in the following we remove the dependency on $k$, even if all the quantities should be intended as specifically computed at the beginning of epoch $k$.

In order to bound the empirical error of the moment estimators, we need to consider the properties of the Markov chain generated by policy $\wt\pi$ and the fact that a single continuous trajectory is observed. In particular, we have to carefully consider the mixing time of the underlying Markov chain (the amount of time it takes that the underlying Markov chain converges to its stationary distribution) and exploit the martingale property of the trajectory. This problem has been previously studied by~\citet{azizzadenesheli2016reinforcement} for the general case of partially observable MDPs. Since \richmdps are a special case of POMDPs, we can directly rely on the following general concentration inequality.

For any ergodic Markov chain with stationary distribution $\omega_{\wt\pi}$, let $f_{1\rightarrow t}(x_t|x_1)$ by the distribution over states reached by a policy $\wt\pi$ after $t$ steps starting from an initial state $x_1$. The inverse mixing time $\rho_{\text{mix},\pi}(t)$ of the chain is defined as
\begin{align*}
\rho_{\text{mix},\wt\pi}(t)=\sup_{x_1}\left\|f_{1\rightarrow t}(\cdot|x_1)-\omega_{\wt\pi}\right\|_{\text{TV}},
\end{align*}
where $\| \cdot\|_{\text{TV}}$ is the total-variation metric. \citet{kontorovich2014uniform} show that for any ergodic Markov chain the mixing time can be bounded as
\begin{align*}
\rho_{\text{mix},\wt\pi}(t)\leq G(\wt\pi)\theta^{t-1}(\wt\pi),
\end{align*}
where $1\leq G(\wt\pi)<\infty$ is the \textit{geometric ergodicity} and $0\leq \theta(\wt\pi)<1$ is the \textit{contraction coefficient} of the Markov chain generated by policy $\wt\pi$.

\begin{proposition}[Theorem 11 \cite{azizzadenesheli2016reinforcement} POMDP concentration bound]\label{lem:POMDPconstration}
Consider a sequence of $\nu$ observations $\{y_1,\ldots,y_\nu\}$ obtained by executing a policy $\wt\pi$ in a POMDP starting from an arbitrary initial hidden state. For any action $l\in[A]$, $\nu^{(l)}$-length sequence $b^{(l)} = \{(y_{t-1},y_{t},y_{t+1});a_t=l\}$, and any $c$-Lipschitz\footnote{under the Hamming metric} matrix valued function $\Phi(\cdot):b^{(l)}\rightarrow \mathbb{R}^{Y\times Y}$, we have
 
%
\begin{align*}
\big\|\Phi(b^{(l)})-\mathbb{E}[\Phi(b^{(l)})]\big\|_2\leq \frac{G(\wt\pi)}{1-\theta(\wt\pi)} \Big(1+\frac{1}{\sqrt{2}c(\nu^{(l)})^\frac{3}{2}}\Big)\sqrt{8c^2\nu^{(l)}\log\Big(\frac{2Y}{\delta}\Big)}
\end{align*}
with probability at least $1-\delta$, where $G(\wt\pi)$ and $\theta(\wt\pi)$ are, respectively, the geometric ergodicity and the contraction coefficient of the underlying Markov chain on the hidden states (they define how fast the underlying Markov chain converges to its stationary distribution), and the expectation is with respect to  the distribution of initial state equals to the stationary distribution. 
\end{proposition}

The parameters $1\leq G(\pi^{(k)})<\infty$ and $0\leq\theta(\pi^{(k)})<1$ are well defined for Markov chain and shows the state distribution of the Markov chain convergence to its stationary distribution (if such distribution exists) with rate of $G(\pi)\theta(\pi)^t$. (the lower $G(\pi)$ and $\theta(\pi)$ give the lower mixing for corresponding Markov chain.) In this paper, we are interested in moments of our data, therefore, $\Phi(\cdot)$ is considered as a moment estimator. 


Given the ergodicity assumption (Asm.~\ref{asm:ergodicity}) under any policy, we can apply Proposition~\ref{lem:POMDPconstration} to bound the errors for both second and third order moments. For any $\lbrace p,q,r\rbrace$ a permutation of set $\lbrace 1,2,3 \rbrace$
\begin{align}\label{eq:perterbationM_2}
\|\wh K_{p,q}^{(l)}-K^{(l)}_{p,q}\|_2&\leq  G(\pi)\frac{1+\frac{1}{\sqrt{2}c(\nu^{(k)}(l))^\frac{3}{2}}}{1-\theta	}\sqrt{8c^2\nu^{(k)}(l)log(\frac{2Y}{\delta})}\\
\|\wh M_{p,q,r}^{(l)}-M_{p,q,r}^{(l)}\|_2&\leq  G(\pi)\frac{1+\frac{1}{\sqrt{2}c(\nu^{(k)}(l))^\frac{3}{2}}}{1-\theta	}\sqrt{8c^2\nu^{(k)}(l)log(\frac{2Y^{1.5}}{\delta})}
\end{align}
with probability at least $1-\delta$. At this point we can proceed with applying the robust tensor power method proposed in~\citep{anandkumar2012method} to recover $V_2^{(l)}$ and obtain the guarantees of Lemma 5 of~\citet{azizzadenesheli2016reinforcement} through Proposition~\ref{lem:concentration}, where $c=\frac{1}{\nu^{(k)}(l)}$. We report a more detailed version of the statement of Lemma~\ref{lem:decomp}.

\begin{lemma}[Concentration Bounds]\label{lem:concentration}
The robust power method of~\citet{anandkumar2012method} applied to tensor $\wh M_3^{(l)}$ returns the $X^{(l)}$ columns of matrix $V_2^{(l)}$ with the following confidence bounds
\begin{align}\label{eq:factor}
\Big\|[V_2^{(l)}]_{(\cdot|i)}-[\widehat{V}^{(l)}_2]_{(\cdot|i)}\Big\|_2\leq \epsilon_3^{(l)}=C^{l}_O\sqrt{\frac{\log(2Y^{3/2}/\delta)}{\nu^{(l)}}}:=\BO^{(l)}
\end{align}
if
\begin{align}\label{eq:nl.condition1}
\nu^{(l)}&\geq\overline{N}:=\max_\pi\left(\frac{4}{{\omega^{(l)}_{\wt\pi_{\min}}\min\limits_{m\in\{1,2,3\}}\lbrace\sigma^2_{\min}(V^{(l)}_m)\rbrace}}\right)^2\log(2\frac{(Y^{1.5})}{\delta})\Theta^{(l)}\\
 \Theta^{(l)}&:=\max\left\{\frac{16(X^{(l)})^{\frac{1}{3}}}{C_1^{\frac{2}{3}}(\omega^{(l)}_{\wt\pi_{\min}})^{\frac{1}{3}}} ,4,\frac{2\sqrt{2}X^{(l)}}{C_1^2\omega^{(l)}_{\min}\min\limits_{m\in\{1,2,3\}}\lbrace\sigma^2_{\min}(V^{(l)}_m)\rbrace}\right\},
\end{align}
with probability at least $1-\delta$, where $C_1$ is a problem-independent constants and $\omega^{(l)}_{\wt\pi_{\min}}:=\min_{i\in\X^{(l)}}\Prob_{\wt\pi}(x=i|a=l)$ where the minimization is over non-zero probabilities. In addition, the $\sigma_{\min}(\cdot)$ operator returns the smallest non-zero singular value of its input matrix. The values of the error $\BO^{(l)}$ under policy $\wt\pi$ is defined (see Eq.29 of~\citet{azizzadenesheli2016reinforcement})
\begin{equation}
\label{eq:epst1}
\BO^{(l)}:=G(\wt\pi)\frac{4\sqrt{2}+2}{(\omega^{(l)}_{\wt\pi_{\min}})^{\frac{1}{2}}(1-\theta(\wt\pi))}\sqrt{\frac{\log(2\frac{(2Y)}{\delta})}{\nu^{(l)}}}+\frac{8\widetilde{\epsilon}^{(l)}}{\omega^{(l)}_{\wt\pi_{\min}}},
\end{equation}
where
\begin{align*}
\widetilde{\epsilon}^{(l)} \leq \frac{2\sqrt{2} G(\wt\pi)\frac{2\sqrt{2}+1}{1-\theta(\wt\pi)}\sqrt{\frac{\log(\frac{2(Y^{\frac{3}{2}})}{\delta})}{\nu^{(l)}}}}{((\omega^{(l)}_{\wt\pi_{\min}})^{\frac{1}{2}}\min\limits_{m\in\{1,2,3\}}\lbrace\sigma_{\min}(V^{(l)}_m)\rbrace)^3}+\frac{\left(64 G(\wt\pi)\frac{2\sqrt{2}+1}{1-\theta(\wt\pi)}\right)}{{\min\limits_{m\in\{1,2,3\}}\lbrace\sigma^2_{\min}(V^{(l)}_m)\rbrace}(\omega^{(l)}_{\wt\pi_{\min}})^{1.5}}\sqrt{\frac{\log\Big(2\frac{Y^{\frac{3}{2}}}{\delta}\Big)}{\nu^{(l)}}},
\end{align*}
\end{lemma}

We notice that the columns of $V_2^{(l)}$ are all orthogonal (but not orthonormal) since the clusters are non-overlapping and an observation $j$ that can be obtained from a state $i$ cannot be generated by any other state $i'$ (i.e., for any $i\neq i'$, $[V_2^{(l)}]_{:,i}^\transp [V_2^{(l)}]_{:,i'} = 0$). As a result, Eq.~\ref{eq:second.moment} can be seen as an eigendecomposition of $M_2^{(l)}$, where the columns $[V_2^{(l)}]_{:,i}$ are the eigenvectors and $\omega_\pi^{(l)}(i)$ are the eigenvalues. More formally, let $M_2^{(l)} = U\Sigma U^{\transp}$ be the eigendecomposition of $M_2^{(l)}$, if all eigenvalues are distinct, the eigenvectors in $U$ can be used to recover $V_2^{(l)}$ up to a mapping function and multiplicative factors.
Nonetheless, in general $V_2^{(l)}$ may have eigenvalues with multiplicity and the eigendecomposition of $M_2^{(l)}$ may return a wrong clustering
 since observations generated by distinct states (and thus with different rewards and dynamics) may be aggregated together. In this case, we have to move to the third order statistics to disambiguate between observations and cluster them properly. 
 
 The computational complexity of this methods appeared at \citep{wang2015fast}, and \citep{song2013nonparametric}.
\section{Rank recovery}\label{sec:rank}
Lemma~\ref{lem:concentration} holds when the rank of matrix $V_2^{(l)}$ is known in advance. While this is not the case in practice, here we show how one can estimate the rank $r = \big|\X^{(l)}_{\pi^{(k)}}\big|$ of $V_2^{(l)}$. Given the expansiveness of latent MDP (Asm.~\ref{asm:expansive}), we have that for any policy $\pi$ and any action $l$, $\big|\X^{(l)}_{\pi}\big|\leq\big|\wb\X^{(l)}_{\pi}\big|$. The rank of the second moment matrix $K^{(l)}_{2,3}$ is then $\min\lbrace \big|\X^{(l)}_{\pi^{(k)}}\big|,\big|\wb\X^{(l)}_{\pi^{(k)}}\big| \rbrace = r$, which also corresponds to the number of non-zero columns in matrix $V_2^{(l)}$. We can then try to estimate $r$ through the estimate second moment $\wh{K}^{(l)}_{2,3}$, which according to Eq.~\ref{eq:perterbationM_2}, estimates $K^{(l)}_{2,3}$ up to an additive error $\epsilon^{(l)}_{2,3}$ that decreases as $O(\sqrt{\frac{1}{\nu^{(l)}}})$. This means that the highest perturbation over its singular values is also at most $O(\sqrt{\frac{1}{\nu^{(l)}}})$.
We introduce a threshold function $g^\epsilon(\nu^{(l)})$ that satisfies the condition
\begin{align}\label{eq:rankcondition}
\epsilon^{(l)}_{2,3} \leq g^\epsilon(\nu^{(l)})\leq 0.5\sigma_{r},
\end{align}
where $\sigma_{r}$ is the smallest non-zero singular value of $K^{(l)}_{2,3}$. We then perform a SVD of $\wh K^{(l)}_{2,3}$ and discard all singular values with value below the threshold $g^\epsilon(\nu^{(l)})$. Therefore, with probability at least $1-\delta$, the number of remaining singular values is equal to the true rank $r$. We are left with finding a suitable definition for the threshold function $g^\epsilon$. From the condition on Eq.~\ref{eq:rankcondition}, we notice that we need $g^{\epsilon}$ to be smaller than a fixed value (RHS) and, at the same time, greater than a decreasing function of order $\mathcal{O}(\sqrt{\frac{1}{\nu^{(l)}}})$ (LHS). Then it is natural to define
\begin{align*}
g^\epsilon(\nu^{(k)}(l))=\frac{g}{\nu^{(k)}(l)^{0.5-\epsilon}}
\end{align*}
for a suitable $g>0$ and with $0<\epsilon<0.5$. Therefore there is a number $N_0^{(l)}$ such that for all $\nu^{(l)}\geq N_0^{(l)}$ the condition on Eq.~\ref{eq:rankcondition} is satisfied and Lemma~\ref{lem:concentration} holds. Therefore we restate the sample complexity in Lemma~\ref{lem:concentration} by adding the extra term to
\begin{align*}
\overline{N}\leftarrow \overline{N} +  N_0(l).
\end{align*} 
Let $\overline{N}_{\max}$ denotes the maximum of this threshold for any action and policy.


\section{Proof of Lemma \ref{lem:partial.cluster}}
Under policy $\pi$, Fig.~\ref{fig:observation.policy}\textit{-right} shows the structure of $V_2^{(l)}$. Given action $l$, the matrix $V_2^{(l)}$ contains $X_{\pi}^{(l)}$ columns and each column corresponds to a column in emission matrix (up to permutation). We showed that the knowledge about a column of $V_2^{(l)}$ reveals part of the corresponding column in emission matrix, the entries with non-zero $\pi(y|l)$. The policy, in general, partitions the observation space to at most $A$ partitions, $\Y_l\forall l\in\A$ and maps each partition to an action. It means that when we condition on an action, e.g., $l$, we restrict ourselves to the part of observation space $\Y_l$ and the input to the spectral learning algorithm is set $\Y_l$. Therefore, the algorithm is able to partition this set to $X_{\pi}^{(l)}$ partition. Because of the unknown permutation over columns of $V_2^{(l)}$ for different actions, we are not able to combine the resulting clustering give different actions. If we enumerate over actions, we end up with $A$ partition $\Y_l$ and then we partition each set $\Y_l$ to at most $X$(upper bound on $X_{\pi}^{(l)}$), as a consequence, we might end up with at most $XA$ disjoint clusters. 

\section{Proof of Lemma \ref{lem:tensor.decomposition}}
We first study the eigendecomposition of $M_2^{(l)}$ when its eigenvalues have multiplicity 1.

\begin{lemma}\label{lem:matrix.decomposition}
For any action $l\in[A]$, let the second moment in Eq.~\ref{eq:second.moment} have the eigendecomposition $M_2^{(l)} = U\Sigma U^{\transp}$. If all eigenvalues of $M_2^{(l)}$ have multiplicity 1, there exists a mapping $\sigma^{(l)}:X\rightarrow X$ and multiplicative constants $\{C_i^{(l)}\}_{i\in[X]}$, such that for any $i\in\X_\pi^{(l)}$ and $j\in[Y]$, $[V_2^{(l)}]_{j,\sigma^{(l)}(i)} = C^{(l)}_i[U]_{j,i}$. As a result, for any hidden state $i\in\X_\pi^{(l)}$ we define the cluster $\wt{\Y}_i^{(l)}$ as
\begin{align}\label{eq:estimated.cluster1}
\wt{\Y}_i^{(l)} = \{j\in[Y]: [U]_{j,i} > 0\}
\end{align}
and we have that if $j,j'\in \wt{\Y}_i^{(l)}$ then $j,j'\in \Y^{(l)}_{\sigma(i)}$ (i.e., observations that are clustered together in $\wt{\Y}_i^{(l)}$ are clustered in the original \richmdp).
\end{lemma}

In Eq.~\ref{eq:second.moment} we show that matrix $M_2^{(l)}$ is a symmetric matrix and has the following representation;
\begin{align*}
M_2^{(l)}:= \sum_{i\in\X^{(l)}_{\pi}} \!\!\omega_\pi^{(l)}(i) [V_2^{(l)}]_{:,i} \otimes [V_2^{(l)}]_{:,i}.
\end{align*}
As long as $V_2^{(l)}]_{:,i}$ for $i\in\X^{(l)}_{\pi}$ are orthogonal vectors, this matrix has rank of $X^{(l)}_{\pi}$ with the following eigendecomposition;
\begin{align*}
M_2^{(l)} = U\Sigma U^{\transp}
\end{align*}
where the matrix $U$, up to permutation, is the orthonormal version of $V_2^{(l)}$, and $\Sigma$ is a diagonal matrix of rank $X^{(l)}_{\pi}$ with diagonal entries equal to $\omega_\pi^{(l)}$ multiplied by the normalization factors. As a result we can use the decomposition to directly recover the non zero elements of $V_2^{(l)}$ and the corresponding partial clustering.

Let's consider the $i$'th and $j$'th nonzero diagonal entries of matrix $\Sigma$, $\sigma_i$ and $\sigma_j$,  with eigenvectors of $U_i, U_j$, i.e., $M_2^{(l)}U_i=\sigma_iU_i~,~M_2^{(l)}U_j=\sigma_jU_j$. In the case of no eigengap, i.e., $\sigma_i=\sigma_j$, for any $0\leq\lambda\leq 1$ we have $M_2^{(l)}(\lambda U_i+(1-\lambda)U_j)= \sigma_i (\lambda U_i+(1-\lambda)U_j)=\sigma_j(\lambda U_i+(1-\lambda)U_j)$. Therefore, any direction in the span of $span(U_i,U_j)$ is an eigenvector and the matrix decomposition is not unique, and we can not learn the true $V_2^{(l)}$. We relax this issue by deploying tensor decomposition of higher order moments. 

The proof of Lemma \ref{lem:tensor.decomposition} directly follows from the properties of tensor decomposition in~\cite{anandkumar2014tensor} and the use of $V_2^{(l)}$ to generate a partial clustering.

%% file: 6Supplementary_regret.tex

\section{Proof of Theorem~\ref{thm:regret}}

The overall proof is mostly based on the original UCRL proof in~\cite{jaksch2010near}. In the following we refer to the exact steps in the original proof whenever we borrow results directly from it. The regret can be decomposed as follows;
\begin{align*}
\textit{Reg}_N=N\eta^*-\sum_{t=1}^Nr_t = \sum_{k=1}^K\underbrace{\sum_{t=t_k}^{t_{k+1}-1} \big(\eta^* - \wb r(x_t,a_t)\big)}_{\Delta_k} + \sum_{t=1}^N \big( \wb r(x_t,a_t) - r_t \big),
\end{align*}
where $K$ is the total (random) number of episodes, $x_t$ is the hidden state of the MDP at time $t$, same as $r_t$ is the reward at time $t$, and $\wb r(x_t,a_t)$ is the true mean of reward. Using Heffting inequallity, as in Eq.~8 in~\cite{jaksch2010near}, the last term can be bounded as $O(\sqrt{N \log(1/\delta)}$ with high probability. We then focus on the per-step regret $\Delta_k$. At any epoch $k$, from Corollary~\ref{cor:partial.cluster.est} we know that any auxiliary state $s\in\wh\S^{(k)}$ is a cluster of observations with same same hidden state. As a result, the reward $\wb r(x_t,a_t)$ is equivalent to $\wb r(y_t,a_t)$ (recall that all observations have the same reward as their hidden state). Therefore;
\begin{align*}
\Delta_k = \sum_{a,s\in\wh\S^{(k)}} \nu^{(k)}(s,a) \big(\eta^* - r(s,a)\big).
\end{align*}
The case when confidence intervals fail is bounded as in the original analysis. We now proceed with the same decomposition as done in~\cite{jaksch2010near} (Eqs.10,13,16) and obtain\footnote{Here we ignore the additive regret coming from approximate extended value iteration that accounts for an extra $O(\sqrt{N})$ regret at the end.}
\begin{align*}
\Delta_k = \bm{\nu}^{(k)} \big( \wt{P}^{(k)} - P^{(k)} \big)w^{(k)} + \sum_{s\in\wh\S^{(k)},a} \nu^{(k)}(s,a)\big( \wt{r}^{(k)}(s,a) - \wb r(s,a) \big) + \bm{\nu}^{(k)} \big( I - P^{(k)} \big)w^{(k)},
\end{align*}
where $\bm{\nu}^{(k)}$ is the vector of number of samples to auxiliary states in epoch $k$, $P^{(k)}$ (resp. $\wt{P}^{(k)}$) is the true (resp. optimistic) transition matrix over auxiliary states of policy $\pi^{(k)}$, $w^{(k)}$ is the centered version of the bias function returned by extended value iteration and $\wt{r}^{(k)}(s,a)$ is the optimistic reward. The first two terms account for the errors in estimating the dynamics and rewards of the (auxiliary) MDP and can be bounded as
\begin{align*}
\Delta_k \leq \square D_{\wh\S^{(k)}} \sqrt{\wh S^{(k)} \log(1/\delta)} \sum_{s\in\wh\S^{(k)},a} \sqrt{N^{(k)}(s,a)} + \bm{\nu}^{(k)} \big( I - P^{(k)} \big)w^{(k)},
\end{align*}
where $D_{\wh\S^{(k)}}$ is the diameter of the auxiliary MDP at epoch $k$ and $\square$ denotes universal numerical constants. The remaining term can be cumulatively bounded following similar steps as in Eq.18 in~\cite{jaksch2010near} with the only difference that the range of $w^{(k)}$ changes at each epoch. Thus we have
\begin{align*}
\sum_{k=1}^K \bm{\nu}^{(k)} \big( I - P^{(k)} \big)w^{(k)} \leq \square \sqrt{\sum_{k=1}^K D_{\wh\S^{(k)}} \nu^{(k)}},
\end{align*}
where $\nu^{(k)} = \sum_{s,a} \nu^{(k)}(s,a)$ is the length of epoch $k$.
Grouping all the terms lead to the first regret statement
\begin{align*}
\textit{Reg}_N \leq \square\bigg(\sum_{k=1}^K D_{\wh\S^{(k)}} \sqrt{\wh S^{(k)} \log(1/\delta)} \sum_{s\in\wh\S^{(k)},a} \sqrt{N^{(k)}(s,a)} + \sqrt{\sum_{k=1}^K D_{\wh\S^{(k)}} \nu^{(k)}}\bigg).
\end{align*}

\subsection{Clustering Rate}\label{app:clustering.rate}

The first regret bound still contains random quantities in terms of the auxiliary MDPs generated over episodes. In this section we derive bounds on the number of steps needed to cluster observations. We notice that the analysis is extremely ``pessimistic'' and as we take worst-case values for all the quantities involved in the analysis.

\textbf{Time to clustering.} We proceed as follows. We first compute the minimum number of samples $\wb N(y)$ to guarantee that an observation $y$ is correctly clustered. We then compute the length $\wb \nu(y)$ of an epoch so that $\wb N(y)$ samples are collected. Finally, we derive how many epochs $\wb K(y)$ are needed before an epoch of length $\wb \nu(y)$ is run. 

We start by defining the probability that a certain action is explored. Let $\pi$ be an arbitrary policy such that action $a$ is taken in at least one observation $y$ belonging to a hidden state $x=x_y$. Whenever an agent is in state $x$, there is a probability $\Prob(y|x)$ to observe $y$ and thus trigger action $a$. We define the probability of ``observing'' an action $a$ in state $x$ under policy $\pi$ as
\begin{align}\label{eq:prob.action}
\alpha_\pi(l) = \sum_{y\in x} \Prob(y|x) \mathbbm{1}(\pi(y)=a).
\end{align}
Since we assumed that $l$ is taken in at least one observation $\alpha_\pi(l)$ is always non-zero and it indeed lower-bounded by $\Omin$. We define $\Pportion:=\min_{k\in[K],x\in\mathcal{X},a\in\mathcal{A}} \alpha_{\pi^{(k)}}(a)$ as the worst proportion across all epochs, states, and actions.

Now we need need to know how fast the set $\mathcal{ S}^{(k)}$ converges to set $\mathcal{X}$, in other work how fast is the clustering process. From Eq.~\ref{eq:BO} and the clustering process, we know that an observation $y$ is clustered in $x_y$ if the number of samples $\nu^{(k)}(l)$ obtained from the action $l$ executed in $y$ within a given epoch $k$ is such that
\begin{align}\label{eq:clustering.condition}
f_O(y|x(y))\geq 2C^{(k)}_O(l)\sqrt{\frac{\log(1/\delta)}{\nu^{k}(l)}}.
\end{align}
By reverting the bound and taking the worse case over actions and epochs, we obtain that a sufficient condition is to collect at least $\wb N(y)$ samples, with
\begin{align*}
\wb N(y) := \max_{l,k} 4C^{(k)}_O(l)\frac{\log(1/\delta)}{(f_O(y|x(y)))^2}.
\end{align*}
We can now leverage on the ergodicity of the MDP and the probability of observation of an action $\Pportion$ to find the minimum number of steps with an epoch to guarantee that with high probability the condition in Eq.~\ref{eq:clustering.condition} is satisfied. We define the worst-case mean returning time as $\tau_M=\max_\pi\max_x \mathbb{E}[\tau_\pi(x\rightarrow x)]$, where $\tau_\pi(x\rightarrow x)$ is the random time to go from $x$ back to $x$ through policy $\pi$.
By Markov inequality, the probability that it takes more than $2\tau_M$ time step to from first visit of state $x$ to its second visit is at most $1/2$. Given the definition of $\Pportion$, it is clear that if the action $l$ is taken in state $x$ then, this action will be taken at state $x$ for $\Pportion$ portion of the time. If we divide the episode of length $\nu$ into $\nu\Pportion/2\tau_M$ intervals of length $2\tau_M/\Pportion$, we have that within each interval we have a probability of $1/2$ to observe a sample from state $x$ and take a particular action. Therefore, the lower bound on the average number of time that the agent takes any action $l$ (that has a non zero probability to be executed in a state $x$) is $\nu\Pportion/4\tau_M$ samples. Thus from Chernoff-Hoeffding, we obtain that the number of samples of any feasible action in the epoch with length $\nu$ is as follows;
\begin{align*}
\forall x\in \mathcal{X},\forall l\in range\lbrace\wt{\pi}(\cdot|x)\rbrace;~\nu(l)\geq \frac{\nu\Pportion}{4\tau_M}-\sqrt{\frac{\nu\Pportion\log(XA/\delta)}{2\tau_M}}
\end{align*}
with probability at least $1-\delta$.
At this point, we can derive a lower bound on the length of the episode that guarantee the desired number of samples to reveal the identity of any observation is reached. For observation $y$, we solve
\begin{align*}
\frac{\nu\Pportion}{4\tau_M}-\sqrt{\frac{\nu\Pportion\log(XA/\delta)}{2\tau_M}}\geq \overline{N}(y)
\end{align*}
and we obtain the condition
\begin{align*}
\sqrt{\nu} \geq \sqrt{\frac{2\tau_{M}}{\Pportion} \log(XA/\delta)} + \sqrt{\frac{2\tau_{M}}{\Pportion} \log(XA/\delta) + \frac{4\tau_{M}}{\Pportion}\overline{N}(y)},
\end{align*}
which can be simplified to
\begin{align}\label{eq:v.condition}
\nu \geq \overline{\nu}(y) := \frac{4\tau_{M}}{\Pportion} \left(\overline{N}(y)+\log(XA/\delta)\right).
\end{align}
With the same argument in App.~D in~\cite{azizzadenesheli2016reinforcement} the number of required epochs to reveal observation $y$ is $\wb{K}(y)\leq AY\log_2(\overline{\nu}(y))+1$. 

\textbf{Time to clustering.} Let $y_{\first} = \arg\min_{y\in\Y} \wb K(y)$ be the first observation that could be clustered\footnote{This should be intended as the first observation that is clustered in the worst case. In practice, depending on the policy and the randomness in the process, other observation may actually be clustered well before $y_1$.} then we define $K_{\first} = \wb K(y_{\last})$ as the number of episodes and $N_{\first} = 4 AY \wb\nu(y_{\first})$ the total number of steps needed before clustering $y_1$ correctly. Similarly, let $y_{\last} = \arg\max_{y\in\Y}$ and $K_{\last} = \wb K(y_{\last})$ as the number of episodes and $N_{\last} = 4 AY \wb\nu(y_{\last})$ the total number of steps needed before clustering $y_{\last}$ correctly. Since all the other observations will be clustered before $y_{\last}$ we can say that by epoch $K_{\last}$ all observations will be clustered. As discussed in Lem.~\ref{lem:partial.cluster} this does not necessarily correspond to the hidden state $\X$ but it could be an auxiliary space $\S$ with at most $AX$ states.

\textbf{Validity of the bound in Lemma~\ref{lem:decomp}.} We also notice that Lemma~\ref{lem:decomp} requires a minimum number of samples $N_0$ before the concentration inequality on the estimate of $V_2$ holds. Applying a similar reasoning as for the time for clustering, we can derive a bound on the number of episodes $K_{\sm}$ and number of samples $N_{\sm}$ needed before the spectral method actually works (from a theoretical point of view). As a result, a more accurate definition of $K_{\first}$ and $K_{\last}$ (resp. for $N$) should take the maximum between the values derived above and $K_{\sm}$.

%% file: 6.1FurtherClustering.tex
\subsection{Minimal Clustering}\label{app:further.clustering}

The spectral learning algorithm has been shown to efficiently cluster the observation set to an auxiliary state space of size $X\leq S\leq XA$. As long as different clusters are merged across epochs, we expect $\S^{(k)}$ to tends to $\X$, yet there is a chance that it converges to a number of auxiliary states $S\neq X$. To make sure that the algorithm eventually converges to the hidden space $\X$, we include a further clustering technique.
We adapt the idea of~\citet{gentile2014online}, \citet{cesa2013gang} and the state aggregation analysis of~\citet{ortner2013adaptive} and perform an additional step of \textit{Reward and Transition Clustering}. In order to simplify the notation, in the following we remove the dependency on $k$, even if all the quantities should be intended as specifically computed at the beginning of epoch $k$. 

We first recall that given any hidden state $x$ and any action $a$ we have $r(y,a) = r(x,a)$ (\textit{reward similarity}) and $p(\cdot|y,a)=p(\cdot|x,a)$ for all observations $y\in\Y_x$ (\textit{transition similarity}). The same similarity measures work for auxiliary states via replacing observations with auxiliary states in the above definitions, i.e., given any hidden state $x$ and any action $a$ we have $r(s,a) = r(s,a)$ and $p(\cdot|s,a)=p(\cdot|x,a)$ for all auxiliary states $s\in\S$ that belong to hidden state $x$.\footnote{Notice that this holds since $\S$ is a ``valid'' clustering in high probability.} We also recall that high-probability confidence intervals can be computed for any $s\in \wh\S$ any $a\in\mathcal{A}$ as
\begin{equation}\label{eq:confidence1}
\begin{split}
&\|p(\cdot|s,a)-\widehat{p}(\cdot|s,a)\|_1\leq d(s,a) := \sqrt{\frac{28\wh S\log(2AN/\delta)}{\max\lbrace 1,N(s,a)\rbrace}}\\
&|\bar{r}(s,a)-\widehat{r}(s,a)|\leq d'(s,a):=\sqrt{\frac{28\log(2\wh SAN/\delta)}{\max\lbrace 1,N(s,a)\rbrace}}.
\end{split}
\end{equation} 
At any epoch, we proceed by merging together all the auxiliary states in $\wh S$ whose reward and transition confidence intervals overlap (i.e., $s$ and $s'$ are merged if the confidence interval $[\wh r(s,a) \pm d_r(s,a)]$ overlaps with $[\wh r(s',a) \pm d_r(s',a)]$ and $[\wh p(\cdot|s,a) \pm d_p(s,a)]$ overlaps with $[\wh p(\cdot|s',a) \pm d_p(s',a)] $) and construct a new set $\wt\S$. In practice, the set $\wt S$ is constructed by building a fully connected graph on $s\in\wh S$ where each state $s$ as a node. The algorithm deletes the edges between the nodes when $|\widehat{r}(s,a)-\widehat{r}(s',a)|>d_r(s,a)+d_r(s',a)$ or $|\widehat{p}(\cdot|s,a)-\widehat{p}(\cdot|s',a)|_1>d_p(s,a)+d_p(s',a)$. The algorithm temporarily aggregates the connected components of the graph and consider each disjoint component as a cluster. If the number of disjoint components is equal to $X$ then it returns $\wh\S$ as the final hidden state $\X$, otherwise the original auxiliary state space $\wh \S$ is preserved and the next epoch is started. Notice that if $s$ and $s'$ belong to the same hidden state $x$ then w.h.p.\ their confidence reward and transition intervals in Eq.~\ref{eq:confidence1} overlap at any epoch. Thus in general $\wt S\leq X$. 

Let's define the reward gaps as follows (similar for the transitions) $\forall s,s'\in\wh \S, \forall a\in\A$ and the corresponding $x,x'$
\begin{align*}
\gamma_r^a(s,s')=\gamma_r^a(x,x'):=\big|\bar{r}(x,a)-\bar{r}(x',a)\big|=\big|\bar{r}(s,a)-\bar{r}(s',a)\big|,\\
\gamma_p^a(s,s')=\gamma_p^a(x,x'):=\|p(\cdot|x,a)-p(\cdot|x',a)\|_1=\|p(\cdot|s,a)-p(\cdot|s',a)\|_1 .
\end{align*}
where $p(\cdot|x,a),p(\cdot|s,a)\in\wt\Delta_{\wh S-1}$, where $\wt\Delta_{\wh S-1}$ is $(\wh S-1)$ dimensional simplex. 
To delete an edge between two states $s,s'$ belonging to two different hidden states, one of the followings needs to be satisfied for at least for one action
\begin{align}
|\widehat{r}(s,a)-\widehat{r}(s',a)|>d_r(s,a)+d_r(s',a)\Rightarrow \gamma_r^a(s,s')>\sqrt{\frac{28\log(2\wh SAN/\delta)}{\max\lbrace 1,N(s,a)\rbrace}}+\sqrt{\frac{28\log(2\wh SAN/\delta)}{\max\lbrace 1,N(s',a)\rbrace}}
\end{align}
\begin{align}\label{eq:rew_bound}
|\widehat{p}(\cdot|s,a)-\widehat{p}(\cdot|s',a)|_1>d_p(s,a)+d_p(s',a)\Rightarrow \gamma_p^a(s,s')  >\sqrt{\frac{28\wh S\log(2AN/\delta)}{\max\lbrace 1,N(s,a)\rbrace}}+\sqrt{\frac{28\wh S\log(2AN/\delta)}{\max\lbrace 1,N(s',a)\rbrace}}
\end{align}
For simplicity, we proceed the analysis with respect to reward, the same analysis holds for transition probabilities. The Eq.~\ref{eq:rew_bound} can be rewritten as follows;
\begin{align*}
\left(\frac{1}{\sqrt{\max\lbrace 1,N(s,a)\rbrace}}+\frac{1}{\sqrt{\max\lbrace 1,N(s',a)\rbrace}}\right)^{-1}\geq\sqrt{\frac{28\log(2\wh SA^2N/\delta)}{\gamma_r^a(s,s')^2}}
\end{align*}
which hold when
 \begin{align}\label{eq:rcluss}
\min\lbrace N(s,a),N(s',a) \rbrace > \frac{112\log(2\wh S AN/\delta)}{\gamma_r^a(s,s')^2}.
 \end{align}

This implies that after enough visits to the auxiliary states $s$ and $s'$, the two states would be split if they belong to different hidden states. We notice that as $\S$ becomes smaller, more and more samples from raw observations are clustered into the auxiliary states, thus making $N(s,a)$ larger and larger. Furthermore, we can expect that the transition gaps may become bigger and bigger as observations are clustered together.

The way that spectral method clusters the observation is effected by separability of observations' probability. But the clustering due to reward analysis (or transition or both) is influenced by the separability in reward function (or transition function or both) and depends on gaps. These two methods look at the clustering problem from different point of view, as a consequence, their combination speeds up the clustering task.

For simplicity we just again look at the reward function, same analysis applies to transition function as well.

\textbf{Regret due to slowness of Reward Clustering (Transition Clustering)}
 Let ${N}^r_a(s,s')$ denote the required number of sample for each of $s$ and $s'$ to disjoint them.
\begin{align*}
{N}^r_a(s,s'):=\frac{56\log(2\wh S AN/\delta)}{\gamma_a(s,s')^2}
\end{align*} 
While the underlying Markov chain is ergodic, with high probability we can say at time step $N(s,s')$, at least for one action $\min\lbrace N(s,a),N(s',a) \rbrace \leq\overline{N}_a(s,s')$ where 
\begin{align*}
N_r(s,s')=\displaystyle\min_a\lbrace\frac{4\tau_{M}}{\Pportion} \left({N}^r_a(s,s')+\log(YA/\delta)\right)\rbrace
\end{align*} 
In the worse case analysis we might need to have $N_r=\max_{s,s'}N_r(s,s')$ samples, which corresponds to at most $AY\log(N_r)$ episode. At this time, the reward clustering procedure can output the exact mapping. This bound can be enhance even further by considering the reward function together with the transition process. With the same procedure we can define $N_p=\max_{s,s'}N_p(s,s')$ where
\begin{align}
N_p(s,s')=\displaystyle\min_a\lbrace\frac{4\tau_{M}}{\Pportion} \left({N}^p_a(s,s')+\log(YA/\delta)\right)\rbrace
\end{align} 
with
\begin{align*}
{N}^p_a(s,s'):=\frac{112\wh S\log(2\wh S AN/\delta)}{\gamma_a(s,s')^2}
\end{align*}
Again, in the worse case analysis we might need to have $N_p=\max_{s,s'}N_p(s,s')$ samples, which corresponds to at most $AY\log(N_r)$ episode.
Therefore the number of required episode for the agent to declare the true mapping w.h.p., is $AY\log (\min \lbrace N_r,N_p\rbrace)$.

%% file: 6.01Consentrations.tex
\section{Concentration inequalities for transition kernel and reward process}\label{Apx:concentration}
In this section we construct the concentration inequality mentioned in Eq.~\ref{eq:confidence1}. 

\textbf{Reward Process}. At first we derive the concentration inequality for the reward process by defining the following martingale sequence at time $\tau$, for a random and incremental sequence of $s^1\subseteq s^2\subseteq \ldots \subseteq s^{\tau}$\footnote{To simplify the notation we use $\S^{\tau'}~,~\tau'\in[1,\ldots]$ to note a random set of clustered which are derived by Alg.~\ref{alg:sm.ucrl} at time $\tau$(we omitted that hat notation for simplicity), and $s^{\tau'}$ as a element in set $\S^{\tau'}$ } (we construct the confidence bound for this cluster as a reference cluster and show the analysis holds for any cluster)
\begin{align*}
\varpi_{\tau}  = \sum_{y\in\Y}\sum_{t=1}^{N}[r_t-\wb r(s^\tau,a)]z_{t,\tau}(y)
\end{align*}
where $z_{t,\tau}(y):=\mathbbm{1}(y_t = y)\mathbbm{1}(y\in s^t) \mathbbm{1}(a_t = a)\mathbbm{1}(t\leq\tau)$
is a function of set $\S^{\tau}$. 
Instead of $\wb r(s^\tau,a)$, we can directly use $\wb r(x,a)$ where $x$ is the corresponding latent state of sequence $s^1\subseteq s^2\subseteq \ldots \subseteq s^{\tau}$. \footnote{Since the regret due to incorrect clustering is considered in the final regret, we assume that the clusters $s^{\tau'}~\tau'\in[1,\ldots]$ are consistent and do not cluster together any two observation which are not from same hidden state.}
We define $N(s,a)$ as a count for state  $s$ and action $a$ at time $N$;
\begin{align*}
N(s,a) = \sum_{y\in\Y}\sum_{t=1}^{N}z_{t,\tau}(y)
\end{align*}
Let's define a Martingale difference for the sequence $\varpi_{\tau}$ as follows:
\begin{align*}
\delta\varpi_{\tau} = \varpi_{\tau} -  \varpi_{\tau-1} = \sum_{y\in\Y}\sum_{t=1}^{N}[r_t-\wb r(s^\tau,a)][z_{t,\tau}(y)-z_{t,\tau-1}(y)]
\end{align*}
Therefore given a filtration $\F$ and  Azuma-Hoeffding inequality \citep{hoeffding1963probability}, since 
\begin{align*}
\sum_{\tau}^N|\delta\varpi|^2\leq N(s,a)~~~~  ,and ~~ \sum_{\tau}^N (Var\lbrace \varpi_{\tau}|\F_{\tau-1}\rbrace)^2\leq N(s,a)^2
\end{align*}
we have the following inequality;
\begin{align*}
\Prob\lbrace \sum_{\tau}^N\delta\varpi_{\tau}\geq \epsilon\rbrace \leq \exp(-\frac{\epsilon^2}{4N(s,a)})
\end{align*}
Therefore we have 
\begin{align*}
\Prob\lbrace |\hat r(s,a)-\wb r(s,a)|\geq \sqrt{\frac{28\log(2 SAN/\delta)}{\max\lbrace 1,N(s,a)\rbrace}}\rbrace\leq \frac{\delta}{60N^7SA}
\end{align*}
Where $\hat r(s,a) := \sum_{y\in\Y}\sum_{t=1}^{N}r_t\mathbbm{1}(y=y_t)z_{t,N}(y)$. Since the $N(s,a)$ is a random variable by its own, we have union bound over time, therefore we have 
\begin{align*}
\Prob\lbrace |\hat r(s,a)-\wb r(s,a)|\geq \sqrt{\frac{28\log(2 SAN/\delta)}{\max\lbrace 1,N(s,a)\rbrace}}\rbrace\leq \sum_n^{N}\frac{\delta}{60N^7SA}<\frac{\delta}{60N^6SA}
\end{align*}

\textbf{Transition Kernel}. Same as before, we need to define a Martingale sequence for transition process. For any possible and fixed clustering $\S$ and a cluster $s$, let's define 
\begin{align*}
\psi_{\tau}  = \sum_{y\in\Y}\sum_{t=1}^{N}[\phi_t-\wb \phi(s,a)]\mathbbm{1}(y\in s) z_{t,\tau}(y)
\end{align*}
where $\phi,\wb \phi\in\Re^S$.
\begin{align*}
[\phi_t]_i= \begin{cases} 1 &\mbox{if } y_{t+1} \in s_i \\ 
0 & \mbox{otherwise} \end{cases},
\end{align*}
$[\wb\phi(s,a)]_i= \Prob(s_i|s,a), \forall i\in S$, $s_i$ is $i$'th cluster in $\S$, and $s^\tau,\tau \in[N]$ is the sequnec of clustering under Alg.~\ref{alg:sm.ucrl}. Given the martingale sequence $\psi_{\tau} $ ( a vector) and the corresponding Martingale difference $\delta\psi_{\tau}:=\psi_{\tau}-\psi_{\tau-1} $ , we construct a new martingale sequnce using dilation technique \cite{tropp2012user}.
\begin{align*}
\Psi_{\tau} = \begin{bmatrix}
       \textbf{0} & \psi_{\tau}            \\[0.3em]
       \psi_{\tau}^\top & \textbf{0}
     \end{bmatrix}
\end{align*}
and 
\begin{align*}
\Delta\Psi_{\tau} = \begin{bmatrix}
       \textbf{0} & \delta\psi_{\tau}            \\[0.3em]
       \delta\psi_{\tau}^\top & \textbf{0}
     \end{bmatrix}
\end{align*}
The corresponding matrices $\Psi_{\tau}$ (and $\Delta\Psi_{\tau}$) is self adjoint matrix which have the following properties;
\begin{align*}
\Psi_{\tau}^2 = \begin{bmatrix}
       \psi_{\tau} \psi_{\tau}^\top &\textbf{0} \\[0.3em]
	   \textbf{0} &\psi_{\tau}^\top \psi_{\tau}
     \end{bmatrix}
\end{align*}
\begin{align*}
\Delta\Psi_{\tau}^2 = \begin{bmatrix}
       \delta\psi_{\tau} \delta\psi_{\tau}^\top &\textbf{0} \\[0.3em]
	   \textbf{0} &\delta\psi_{\tau}^\top \delta\psi_{\tau}
     \end{bmatrix}
\end{align*}
and $\lambda_{max}(\Psi_{\tau})=\|\Psi_{\tau}\|_2=\|\psi_{\tau}\|_2$. Therefore using Freedman inequality \citep{tropp2011freedman}, and the covariance matrix:
\begin{align*}
W_\tau := \sum_{\tau'=1}^{\tau}\E[\Delta\Psi^2_{\tau'}|\F_{\tau'-1}],~\textit{and}, \lambda_{\max}(\Delta\Psi_{\tau})\leq C ~,~\forall \tau
\end{align*}
for all $\epsilon\geq 0$ and $\sigma^2>0$
\begin{align*}
\Prob[\exists \tau'\geq 0:\lambda_{\max}(\Psi_{\tau'})\geq \epsilon~ \textit{and}~ \|W_{\tau'}\|\leq \sigma^2]\leq 2S.\exp\lbrace-\frac{-\epsilon^2}{2\sigma^2+2C t/3}\rbrace
\end{align*} 
in other word
\begin{align*}
\|\Psi_\tau\|_2\leq \frac{2C}{3}+\sqrt{2\sigma^2\log(S/\delta)} 
\end{align*}
with probability at least $1-\delta$. Since $\sigma^2\leq \|W_\tau\|_1\leq N'(s,a)$ where $N'(s,a):=\sum_{y\in\Y}\sum_{t=1}^{N}\mathbbm{1}(y\in s) z_{t,\tau}(y)$, we have

\begin{align*}
\|\psi_\tau\|_2\leq \frac{2C}{3}+\sqrt{2N'(s,a)\log(S/\delta)} 
\end{align*}
and for the average $\alpha_\tau := \psi_\tau/(N'(s,a))$
\begin{align*}
\|\alpha_\tau\|_2\leq \frac{2C}{3N'(s,a)}+\sqrt{\frac{2\log(S/\delta)}{N'(s,a)}} 
\end{align*}
Since $C\leq 2$ and $S\leq Y$, for $N(s,a)\geq 1$
\begin{align*}
\|\alpha_\tau\|_2\leq \sqrt{\frac{4\log(Y/\delta)}{N'(s,a)}} 
\end{align*}
Therefore, if we replace $\delta$ with $\frac{\delta}{20N^7YA}$ we get 

\begin{equation}
\begin{split}
&\|\Prob(\cdot|s,a)-\widehat{\Prob}(\cdot|s,a)\|_1\leq d(s,a) := \sqrt{\frac{28 S\log(2AN/\delta)}{\max\lbrace 1,N'(s,a)\rbrace}}\\
\end{split}
\end{equation} 
Since this inequality holds for any clustering $\S$, it holds for $\S^N$, generated by Alg.~\ref{alg:sm.ucrl} as well. Therefore, at time step $N$, if we are interested in $s^N \in \S^N$, choose $s = s^N$ while $N'(s,a) = N(s^N,a)$, then we have 
\begin{equation}
\begin{split}
&\|\Prob(\cdot|s^N,a)-\widehat{\Prob}(\cdot|s^N,a)\|_1\leq d(s,a) := \sqrt{\frac{28 S^N\log(2AN/\delta)}{\max\lbrace 1,N(s^N,a)\rbrace}}\\
\end{split}
\end{equation}

%% file: 7MoreExperiments.tex
\section{Extended Discussion}\label{Aux:deter}

\begin{conjecture}
We can extend this results to deterministic MDP. In this case we can first uniformly explore the latent space and collect sufficient number of sample to find the exact clustering and reduce the large MDP to the latent MDP and then apply \ucrl on the latent MDP. Which can suffer a constant regret of pure exploration at the beginning and regret of $\mathcal{\wt O}(D_\X X\sqrt{AN})$ due to \ucrl in the second phase.  
One of the main open questions is whether the spectral clustering method could still provide ``useful'' clusterings when the state space is not fully visited (i.e., in case of non-ergodic MDP), so that observations are properly clustered where it is actually needed to learn the optimal policy. We can provide a partial answer in the case of deterministic \richmdps. In fact, despite not being ergodic, in this case we can first uniformly explore the latent space and collect sufficient number of sample to find the exact clustering and reduce the large MDP to the latent MDP and then apply \ucrl on the latent MDP. These two-phase algorithm would suffer a constant regret of pure exploration at the beginning and regret of $\mathcal{\wt O}(D_\X X\sqrt{AN})$ due to \ucrl in the second phase. 
\end{conjecture}
In RL problems, the principle of Optimism-in-Face-of-Uncertainty contributes in designing a policy that locally improves the model uncertainty and average reward which has been shown to be an optimal strategy. It is an open question to analyze and modify this principle for the models with clustering where global improvement of the information in model uncertainty is required. While the $\smcmdp$ for deterministic models reaches order optimal regret, it is not still clear how to modify the exploration to enhance the constant regret of the pure exploration phase.

\textbf{Compared to POMDP}.
We can compare this result with the regret bound of~\citet{azizzadenesheli2016reinforcement} for POMDPs, which are a more general class than \richmdps. Recalling that POMDPs are characterized by a diameter 
\begin{align*}
D_{pomdp}:=\max_{x,x',a,a'}\min_{\pi\in\mathcal{P}} \mathbb{E}[\tau\left(\left(x,a\right)\rightarrow\left(x',a'\right)\right)],
\end{align*}
the regret derived by~\citet{azizzadenesheli2016reinforcement} scales as $\wt{\mathcal{O}}(D_{pomdp}X^{3/2}\sqrt{AYN})$. The regret suffers from additional term $\sqrt{Y}$ because the RL algorithm in POMDP put much effort on accurate estimation of entries of $O$ matrix and does not exploit its specific structure. Moreover, there is an additional factor $X$ in regret bound due to learning of transition tensor through spectral methods.

\section{Additional Experiments}\label{app:experiment}
\begin{figure}[ht]
\centering
  \includegraphics[height=4.5cm]{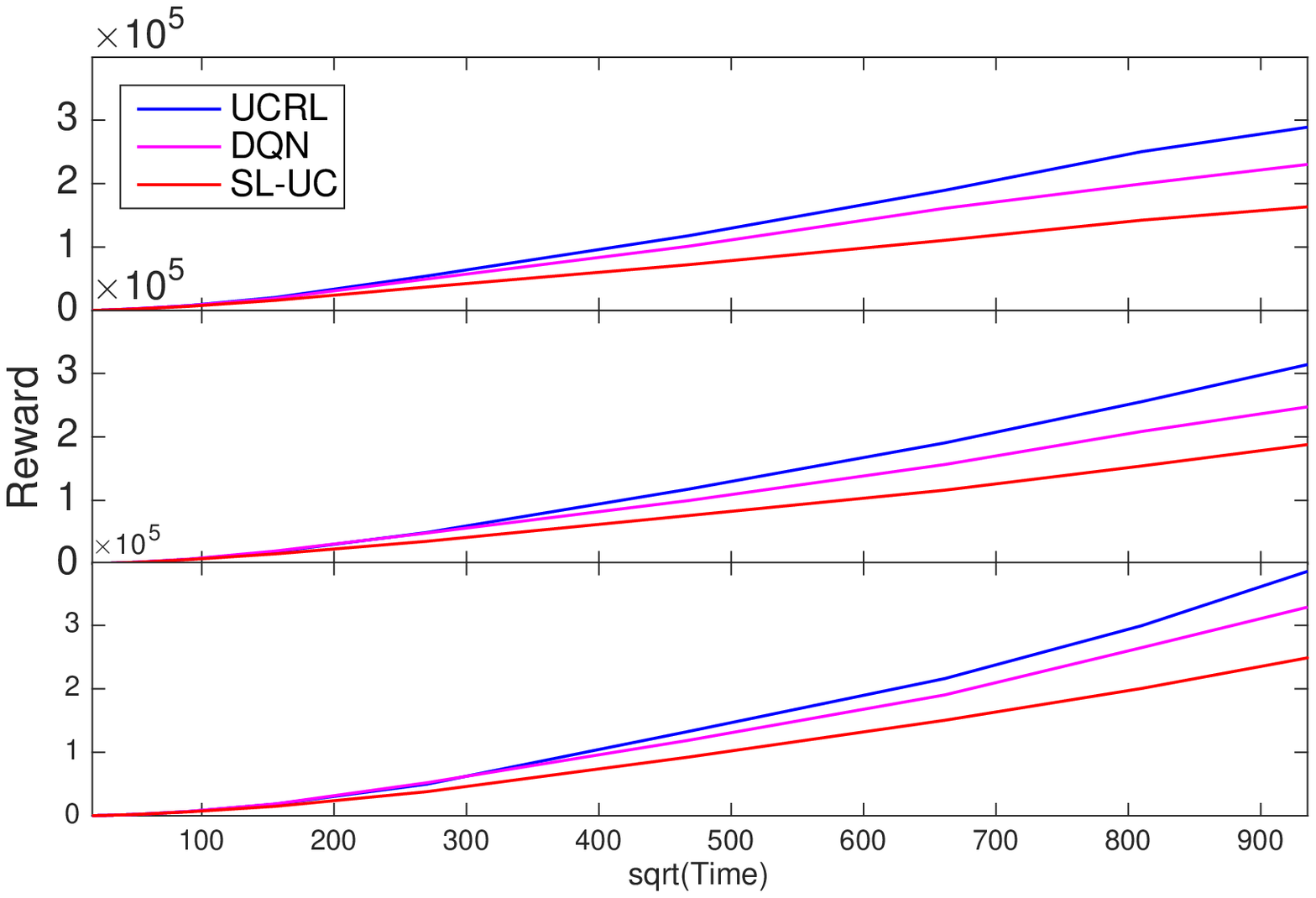}
\hspace{0.5in}
  \centering
  \includegraphics[height=4.5cm]{Fig/Plot/regretcompareMDP.eps}
\caption{\textit{(left)}The regret comparison, $A=4$, from top to bottom, $Y=10,20,30$. The scale is $\sqrt{T}$. \textit{(right)} Learning rate of \smcmdp compared to \ucrl and DQN. After first few rounds, it learns the the true mapping matrix. The numbers in the bulbs are the cardinality of Aux-MDP. }
\label{fig:comparisonCar}
\end{figure}

While the results reported in the main text are obtained on actual \richmdps, here we test \smcmdp on random MDPs with no explicit hidden space. The objective is to verify whether \smcmdp can be used to identify (approximate) clusters. Since \smcmdp in high probability only clusters observations that \textit{actually} belong to the same hidden state, in this case \smcmdp would reduce to run simple UCRL, as there is no two observations that can be \textit{exactly} clustered. In order to encourage clustering, we half the (exact) confidence intervals in the attempt of trading off a small bias with a significant reduction in the variance. We compare the regret on three random MDPs with increasing number of states. As it is shown in Fig.~\ref{fig:comparisonCar}, \smcmdp is effective even in this scenario compared to UCRL and DQN. In fact, we see from Fig.~\ref{fig:comparisonCar}-right that \smcmdp is able to find clusters without compromising the overall regret. While the number of states now directly affects the performance of \smcmdp, we see that it is more robust than the other algorithms and its regret is not severely affected by an increasing number of observations.

